\documentclass{article}
\usepackage{iclr2027_conference,times}

\usepackage{amsmath,amsfonts,bm}

\def\eqref#1{equation~\ref{#1}}

\def\1{\bm{1}}

\DeclareMathAlphabet{\mathsfit}{\encodingdefault}{\sfdefault}{m}{sl}
\SetMathAlphabet{\mathsfit}{bold}{\encodingdefault}{\sfdefault}{bx}{n}

\usepackage{hyperref}
\usepackage{url}
\usepackage{placeins}

\usepackage{amsmath,amssymb}
\usepackage{booktabs}
\usepackage{microtype}
\usepackage{placeins}
\usepackage{float}
\usepackage{array}
\usepackage{tabularx}
\usepackage[table]{xcolor}
\usepackage{graphicx}
\usepackage{multirow}
\usepackage{makecell}
\definecolor{headergray}{gray}{0.90}
\newcolumntype{L}[1]{>{\raggedright\arraybackslash}m{#1}}
\newcolumntype{C}[1]{>{\centering\arraybackslash}m{#1}}
\usepackage{enumitem}
\usepackage{booktabs}
\usepackage{algorithm}
\usepackage{amsmath}
\usepackage{amssymb}
\usepackage{algpseudocode}
\newcolumntype{L}[1]{>{\raggedright\arraybackslash}p{#1}}
\newcolumntype{Y}{>{\raggedright\arraybackslash}X}
\usepackage[most]{tcolorbox}
\newtcblisting{promptbox}[1][]{
    enhanced,
    listing only,
    breakable,
    colback=gray!6,
    colframe=gray!55,
    colbacktitle=gray!35,
    coltitle=black,
    fonttitle=\rmfamily\bfseries\small,
    boxrule=0.5pt,
    arc=1.5pt,
    left=6pt,
    right=6pt,
    top=5pt,
    bottom=5pt,
    listing options={
        basicstyle=\rmfamily\small,
        breaklines=true,
        columns=fullflexible,
        keepspaces=true,
        showstringspaces=false
    },
    #1
}

\title{FIT-MAC: A Closed-Loop Multi-Agent Framework for Robust Vision-Language-Action Models}

\author{
\normalfont
\begin{tabular}{c}
\textit{
Jiatong Li\textsuperscript{1},
Jisheng Dang\textsuperscript{1},
Qiangqiang Ma\textsuperscript{2},
Bimei Wang\textsuperscript{1},
Jingwen Zhao\textsuperscript{3}
}
\\
\textit{
Wenjun Ma\textsuperscript{1},
Juncheng Li\textsuperscript{4},
Hong Peng\textsuperscript{1},
Bin Hu\textsuperscript{5}
}
\\[0.6em]
\small
\textsuperscript{1}Lanzhou University
\quad
\textsuperscript{2}Northwest Normal University
\\
\small
\textsuperscript{3}Sun Yat-sen University
\quad
\textsuperscript{4}Zhejiang University
\quad
\textsuperscript{5}Beijing Institute of Technology
\end{tabular}
}

\iclrfinalcopy

\begin{document}
\maketitle
\lhead{}  

\begin{abstract}
Vision-language-action (VLA) models have achieved strong performance on standard robotic manipulation benchmarks. However, their reliability can degrade substantially when task conditions deviate from the nominal setting. Perturbations in object positions and task instructions may lead to complete task failure. Existing VLA policies generally lack an explicit runtime mechanism for distinguishing these failure modes and selecting an appropriate recovery strategy.
To systematically mitigate this issue, we propose FIT-MAC, a failure-aware, closed-loop multi-agent control framework for improving the robustness of VLA policies at inference time.
FIT-MAC augments a VLA policy with an inference-time control layer. First, the Planner Agent parses the language instruction into a structured task specification that captures the manipulation goal and spatial constraints, and the Grounder Agent resolves each instruction-level reference to a visual target. Next, the Verifier Agent tracks phase-specific execution progress, distinguishes target-binding errors from local execution failures, and determines whether intervention is warranted. The Correction Agent then routes permitted interventions to target rebinding or bounded grasping or placement recovery, depending on the failure type. When no intervention is required, the pretrained VLA remains the default action generator.
Extensive experiments on LIBERO-PRO demonstrate that FIT-MAC improves robustness across multiple VLA backbones, consistently increasing success rates and achieving state-of-the-art performance under diverse perturbations. Code\footnote{Project page: \url{https://fitmac-research.pages.dev/}.} is available at \url{https://anonymous.4open.science/r/research-artifact-B431}.
\end{abstract}

\section{INTRODUCTION}

Vision-language-action (VLA) models integrate visual perception and language understanding to generate actions for robotic manipulation. OpenVLA provides an open-source foundation for generalist robot control \citep{kim2024openvla}, while $\pi_0$ uses flow matching to generate continuous actions \citep{black2025pi0}. Its successor, $\pi_{0.5}$, further improves open-world generalization through heterogeneous co-training \citep{physicalintelligence2025pi05}. Recent models pursue complementary directions, NORA reduces the computational cost of generalist VLA policies \citep{hung2025nora}, while X-VLA targets scalable cross-embodiment transfer \citep{zheng2025xvla}. MolmoAct incorporates structured spatial reasoning between perception and low-level control \citep{lee2025molmoact}. Other methods improve predictive modeling and long-horizon execution through world-knowledge modeling, future-goal prediction, and phase-aware policies \citep{upvla2025,dreamvla2025,longvla2025}. Collectively, these advances highlight the potential of VLA models for generalist robot control.

However, strong benchmark performance does not necessarily translate into reliable execution under distribution shift. AGNOSTOS reports limited zero-shot generalization across tasks, while SAFE identifies frequent failures on novel tasks, underscoring the need for runtime failure detection \citep{agnostos2025,safe2025}. Broader evaluations show substantial sensitivity to perturbations in actions, instructions, observations, and environments \citep{robustvla2026,zhou2026liberopro}. VLATest documents similar robustness gaps under variations in clutter, lighting, viewpoints, and instructions, as well as in the presence of unseen objects \citep{vlatest2025}. Adversarial evaluations further expose vulnerabilities to digital and physical perturbations \citep{advvla2025}. Together, these findings highlight the challenges of robust VLA execution.


Addressing these challenges requires both reliable temporal visual evidence and
effective runtime recovery. Reasoning-based video segmentation improves
language-conditioned grounding by decoupling textual semantics from dynamic
visual representations \citep{dang2026decoupled}. At the control level, RACER
combines a VLM supervisor with an actor trained on automatically generated
recovery trajectories \citep{racer2025}. Sparse spatio-temporal propagation
strengthens object correspondence across frames \citep{dang2026fast}, while
COME-robot uses execution feedback for closed-loop failure diagnosis and
recovery \citep{comerobot2025}. Frame-relation mining reduces redundant memory
and mitigates error accumulation \citep{dang2023efficient}, whereas TCoT
integrates hierarchical reasoning with global--local recovery
\citep{tcot2026}. Parallel temporal--spatial memory further captures
complementary motion and contextual cues \citep{dang2024temporo}. 

\begin{figure}[t]
\centering
\includegraphics[width=\linewidth]{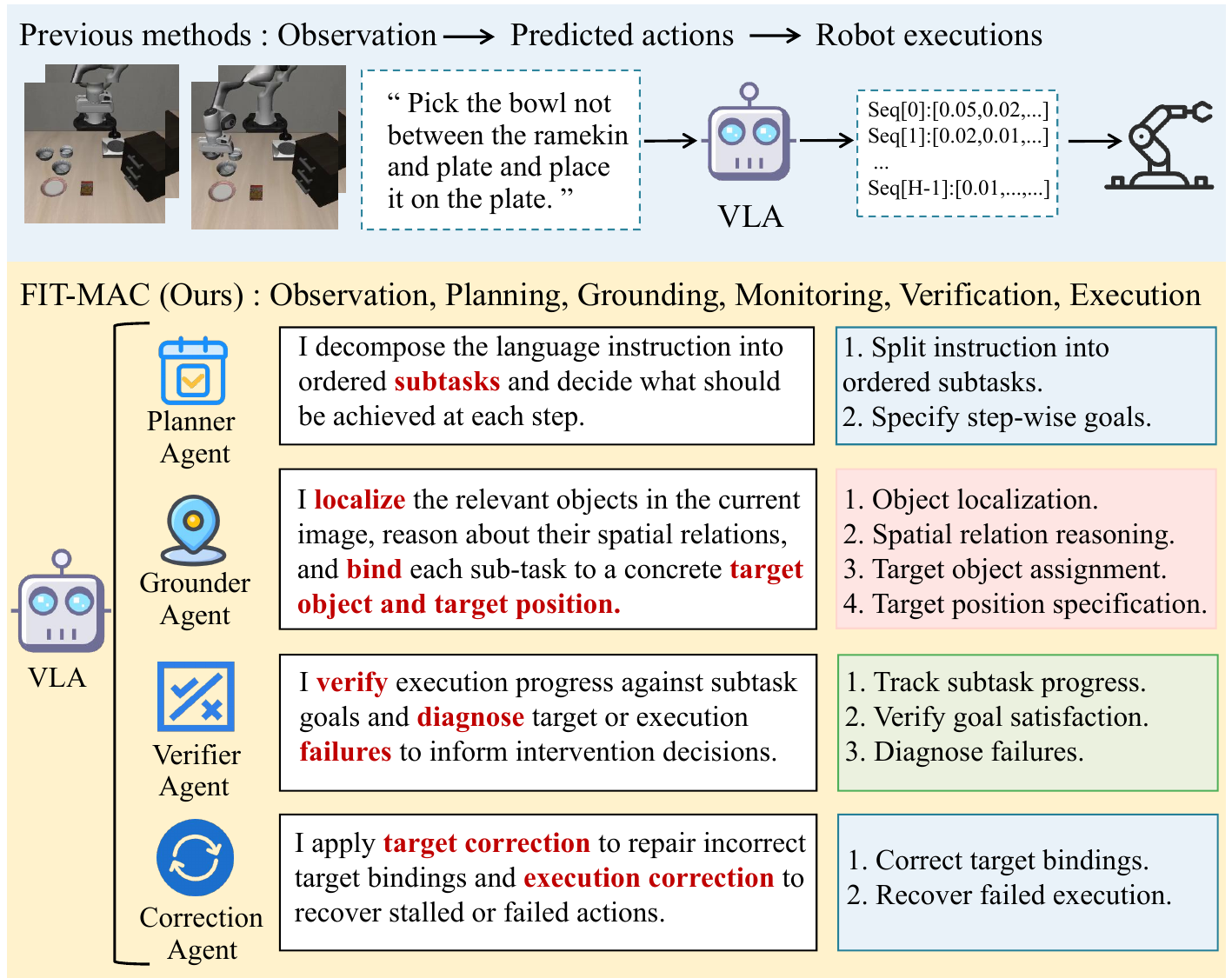}
\vspace{-15pt}
\caption{Comparison of standard VLA execution and FIT-MAC. Top: a VLA policy predicts actions from visual observations and a language instruction. Bottom: FIT-MAC augments the pretrained policy with specialized agents.}
\vspace{-5pt}
\label{fig:motivation}
\end{figure}

We study a complementary runtime decision problem of distinguishing incorrect target binding from local execution failure and selectively applying the appropriate recovery mechanism. The pretrained VLA remains the default action generator.
As illustrated in Figure~\ref{fig:motivation}, we aim to improve the robustness of pretrained VLA policies through runtime recovery without modifying the underlying policy. Our key observation is that perturbations induce distinct failure types. Target failures arise when incorrect grounding of the language-specified target object or reference relation directs the policy toward the wrong object instance before grasping. Execution failures occur when the target is correctly identified,  but local control becomes unstable during grasping, transport, or placement. These failures manifest as empty grasps, failed lifts, unstable releases, or stalled execution. Distinguishing between these failure types allows an external controller to select targeted recovery mechanisms rather than apply generic retries.

Building on this observation, we propose FIT-MAC, a failure-aware, closed-loop multi-agent control framework for improving the robustness of VLA policies at inference time. Specialized agents coordinate task parsing, visual grounding, execution monitoring, and recovery through a shared execution state. FIT-MAC retains the pretrained VLA as the default action generator and intervenes only when task progress, target visibility, or grasping and placement signals indicate failure. It distinguishes incorrect target binding from local execution failure and routes them to target rebinding or bounded local recovery, respectively. This design enables selective, interpretable correction while limiting disruption to nominal execution and leaving the pretrained policy unchanged.

Experiments on LIBERO-Object and LIBERO-Spatial show that FIT-MAC consistently improves the robustness of OpenVLA, $\pi_0$, and $\pi_{0.5}$ across diverse perturbations. The largest gains occur under position, task, and environment perturbations, where successful execution often requires target rebinding and stage-wise recovery. FIT-MAC also maintains high success rates under object and semantic perturbations, demonstrating its effectiveness across perturbation types. Our contributions are as follows:

\begin{itemize}[leftmargin=18pt, itemsep=2pt, topsep=2pt]
  \item We propose FIT-MAC, a failure-aware, closed-loop multi-agent control framework that improves the robustness of VLA policies at inference time. Specialized agents coordinate monitoring, diagnosis, routing, and correction through a shared execution state, while the pretrained VLA remains the default action generator.
  \item We introduce a runtime verification and correction mechanism that assesses target grounding and execution progress to determine when to intervene. It distinguishes target-binding errors from local execution failures and addresses them through target rebinding and bounded local recovery, respectively.
  \item FIT-MAC achieves state-of-the-art success rates across LIBERO-PRO perturbation settings and consistently improves the robustness of OpenVLA, $\pi_0$, and $\pi_{0.5}$. Ablation studies further demonstrate the contribution of each component.
\end{itemize}

\section{RELATED WORK}

\vspace{-3pt}
\paragraph{Generalist and Long-Horizon Vision-Language-Action Models.}
Generalist VLA models learn language-conditioned visuomotor policies from heterogeneous robotic data. OpenVLA provides an open-source autoregressive foundation model \citep{kim2024openvla}. $\pi_0$ uses flow matching for continuous action generation, while $\pi_{0.5}$ improves open-world generalization through heterogeneous co-training \citep{black2025pi0,physicalintelligence2025pi05}. NORA improves inference efficiency, X-VLA supports cross-embodiment transfer through soft prompts, and MolmoAct integrates depth-aware perception, spatial reasoning, and action prediction \citep{hung2025nora,zheng2025xvla,lee2025molmoact}.
Recent approaches address long-horizon manipulation through hierarchical reasoning, memory, future-state prediction, and spatial representations. AtomVLA decomposes demonstrations into atomic subtasks and scores action chunks using a latent world model \citep{sun2026atomvla}. LaMP incorporates a 3D scene-flow prior, R3DP introduces a real-time 3D-aware policy, and RoboTALES uses task-aligned simulated futures to guide policy learning \citep{lamp2026,r3dp2026,robotales2026}. These methods modify the policy architecture, representation, or training procedure.

\paragraph{Perturbation Evaluation and Robust Policy Optimization.}
Benchmarks such as LIBERO-PRO, LIBERO-X, VLATest, adversarial VLA benchmarks, and LIBERO-RECOVER assess VLA robustness to changes in objects, positions, language, observations, and environments \citep{zhou2026liberopro,liberox2026,vlatest2025,advvla2025,liberorecover2026}. Existing methods improve robustness through perturbation-based training, post-training, uncertainty-aware action refinement, or latent world-model scoring \citep{robustvla2026,robustvla_rl2025,islam2026sureflow,sun2026atomvla}. VLS instead uses a vision-language model to steer a pretrained policy at inference time \citep{vls2026}.

\paragraph{Online Failure Detection and Agentic Runtime Recovery.}

Runtime recovery methods monitor execution and intervene when failures occur. SAFE, RecoveryChaining, RACER, COME-robot, and TCoT explore failure detection, recovery to resumable states, VLM supervision, execution tracing, and hierarchical recovery \citep{safe2025,recoverychaining2025,racer2025,comerobot2025,tcot2026}. GTA-VLA uses interactive visual guidance, including points, boxes, and traces, to steer VLA execution under spatial and visual ambiguity \citep{gtavla2026}. Code-based and agentic systems extend runtime recovery through execution feedback, visual verification, skill synthesis, and modular coordination \citep{fu2026capx,zhang2026rats,modularfailure2025,malmm2025,optimizedprompts2025,agenticrobot2025,robobridge2026}.

Our method explicitly distinguishes target failures from execution failures and addresses them through target rebinding with closed-loop correction and bounded local grasping and placement control, respectively. The pretrained VLA remains the default action generator, and intervention is triggered only by runtime evidence of failure.

\section{METHOD}
\label{sec:method}

\begin{figure}[t]
    \centering
    \includegraphics[width=\textwidth]{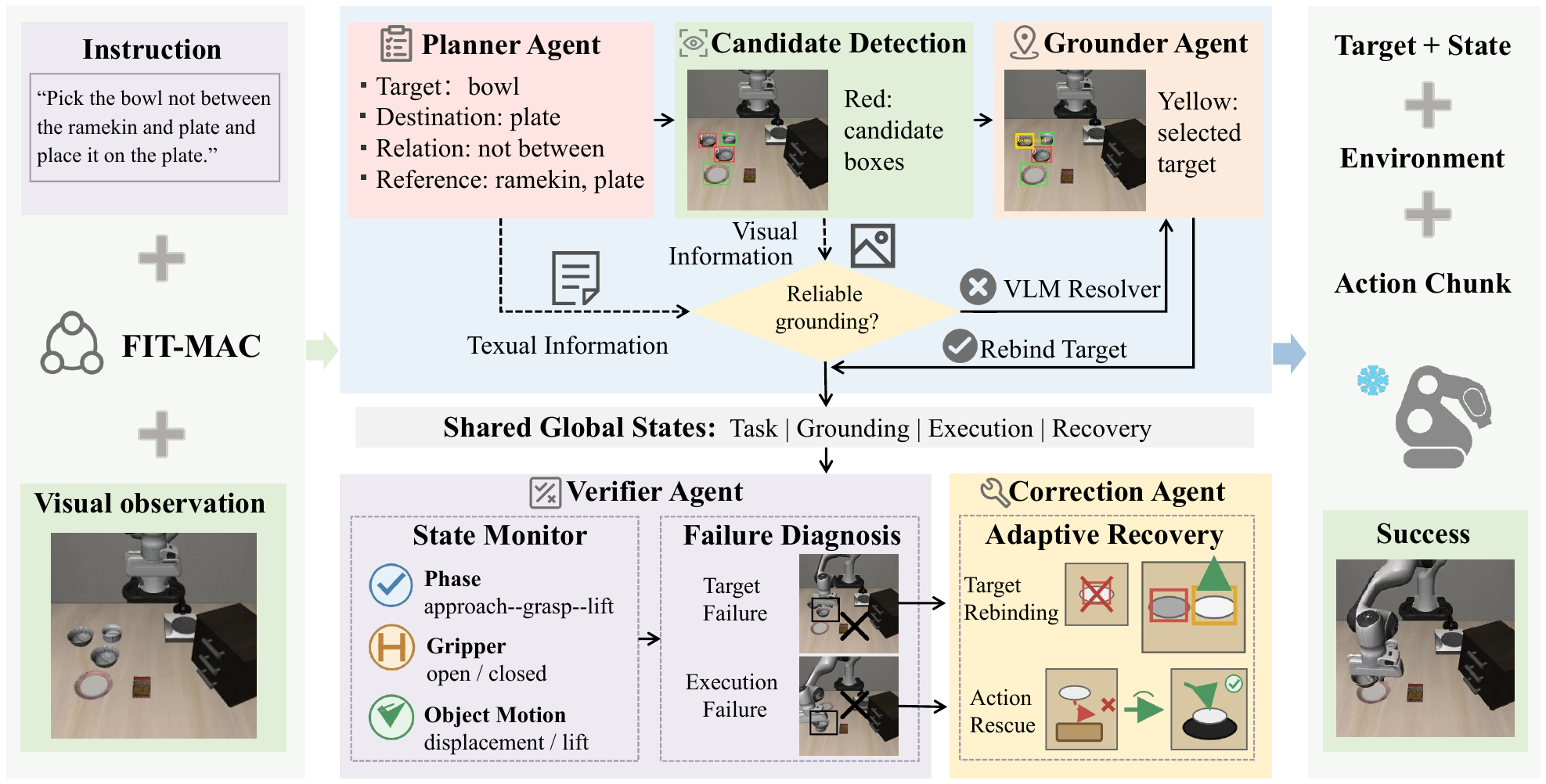}
    \vspace{-15pt}
    \caption{Overview of FIT-MAC. The framework first establishes an instance-level target lock through target rebinding, then monitors stage-specific execution progress and routes detected failures to target correction or bounded local recovery.}
    \label{fig:pipeline}
\end{figure}

\vspace{-5pt}

\subsection{Overview}
\label{sec:method_overview}

In Figure~\ref{fig:pipeline}, we provide an overview of FIT-MAC. Given a language instruction $x$, a visual observation $I_t$, and a low-dimensional robot state $o_t$, FIT-MAC augments a frozen VLA policy $\pi_{\mathrm{VLA}}$ with an inference-time control layer. The framework comprises a Planner Agent, a Grounder Agent, a Verifier Agent, and a Correction Agent that coordinate through a shared execution state. The Planner and Grounder Agents convert the instruction into a grounded task specification, while the Verifier Agent tracks execution progress and determines whether intervention is needed. Target-binding errors trigger target rebinding, whereas local execution failures trigger bounded recovery. The pretrained VLA remains the default action generator throughout execution.

\subsection{Planner Agent}
\label{sec:task_prompt}

The Planner Agent converts the language instruction into a structured task specification that guides grounding and execution. It extracts the target object, source and destination descriptions, spatial relations, and reference objects. For multi-step manipulation instructions, it organizes the required operations into an ordered sequence of subtasks. The resulting specification explicitly defines each subtask's goal and constraints for downstream agents.

Execution proceeds through bounded pick-and-place attempts. Let $k_t$ denote the active subtask index and $\phi_t$ the current manipulation phase. The phase belongs to either the pick-phase set $\Phi_{\mathrm{pick}}$ or the place-phase set $\Phi_{\mathrm{place}}$. Both variables are maintained in the shared execution state, allowing the Verifier and Correction Agents to apply stage-specific criteria.

\subsection{Grounder Agent}
\label{sec:target_correction}

The Grounder Agent maps the structured task specification to object instances in the current observation. Before each subtask, Grounding DINO detects candidate target, destination, and reference objects. The Grounder Agent then matches the detected candidates to the descriptions and spatial relations specified in the instruction. When detections are uncertain or ambiguous, a VLM resolver selects the candidate that best matches the task specification. The selected target and destination are stored in the grounding state $g_t$, which is used to construct the grounded prompt $\tilde{x}_{k_t}$ for the pretrained VLA.

To incorporate geometric evidence into target selection, we score candidates within the instruction-compatible set $\mathcal{C}_t^{\mathrm{sem}}$ using
\begin{equation}
S_t(c)
=
\left\|p_{t,\mathrm{xy}}^{c}
-p_{t,\mathrm{xy}}^{\mathrm{ee}}\right\|_2
+
\lambda_z
\left|p_{t,z}^{c}-p_{t,z}^{\mathrm{ee}}\right|,
\label{eq:geometric_candidate_score}
\end{equation}
where $p_t^c$ and $p_t^{\mathrm{ee}}$ denote the candidate and end-effector positions, respectively, and $\lambda_z$ weights the vertical distance. The lowest-scoring candidate is $\hat{c}_t=\arg\min_{c\in\mathcal{C}_t^{\mathrm{sem}}}S_t(c)$. Restricting this comparison to instruction-compatible candidates ensures that geometric proximity does not override semantic compatibility.

The selected target instance is retained as a target lock throughout the current subtask. The grounding state is shared with the Verifier Agent for consistency checks and the Correction Agent for recovery. When target rebinding is requested, the Grounder Agent updates the binding and grounded prompt.

\subsection{Verifier Agent}
\label{sec:monitor_diagnosis}

The Verifier Agent checks whether the target binding remains reliable and execution is progressing as expected. It monitors target approach, gripper closure, object--gripper motion, destination approach, and post-release stability. These observations help determine whether the robot is acting on the intended instance and whether the current manipulation attempt is making progress.

Target verification assesses the confidence $\kappa_t$ of the active target lock and its consistency with the current observation. The binding requires correction when $\kappa_t<\tau_\kappa$ or $\mathrm{Inconsistent}(g_t,I_t)$ holds. The latter condition indicates that the locked instance no longer matches the observed scene or the task-specified spatial relation.

Execution verification measures the end-effector--target distance during picking and the object--destination distance during placement as
\begin{equation}
d_t^{\mathrm{tgt}}
=
\left\|p_t^{\mathrm{ee}}-p_t^{\mathrm{tgt}}\right\|_2,
\qquad
d_t^{\mathrm{dst}}
=
\left\|p_t^{\mathrm{obj}}-p_t^{\mathrm{dst}}\right\|_2,
\label{eq:stage_distances}
\end{equation}
where $p_t^{\mathrm{tgt}}$, $p_t^{\mathrm{obj}}$, and $p_t^{\mathrm{dst}}$ denote the grounded target, manipulated object, and destination positions, respectively. Progress over $\Delta$ control steps is quantified by the distance reductions
\begin{equation}
\Delta d_t^{\mathrm{tgt}}
=
d_{t-\Delta}^{\mathrm{tgt}}-d_t^{\mathrm{tgt}},
\qquad
\Delta d_t^{\mathrm{dst}}
=
d_{t-\Delta}^{\mathrm{dst}}-d_t^{\mathrm{dst}}.
\label{eq:stage_progress}
\end{equation}
Positive values indicate reduced distance to the target object or destination, respectively. Gripper closure, object--gripper motion, and post-release stability provide additional evidence for assessing grasping, lifting, and placement completion.

We combine distance reduction and observed motion into a phase-specific progress indicator defined as
\begin{equation}
\chi_t^{\mathrm{prog}}
=
\begin{cases}
\mathbf{1}\!\left[
\Delta d_t^{\mathrm{tgt}}>\tau_{\mathrm{tgt}}
\ \lor\
\|\Delta p_t^{\mathrm{ee}}\|_2>\tau_{\mathrm{ee}}
\right],
& \phi_t\in\Phi_{\mathrm{pick}},\\[2mm]
\mathbf{1}\!\left[
\Delta d_t^{\mathrm{dst}}>\tau_{\mathrm{dst}}
\ \lor\
\|\Delta p_t^{\mathrm{obj}}\|_2>\tau_{\mathrm{obj}}
\right],
& \phi_t\in\Phi_{\mathrm{place}}.
\end{cases}
\label{eq:progress_indicator}
\end{equation}
Here, $\Delta p_t^{\mathrm{ee}}$ and $\Delta p_t^{\mathrm{obj}}$ denote the corresponding displacements over $\Delta$ control steps. The thresholds specify the minimum distance reduction or displacement counted as progress.

A phase is considered stalled if no progress is detected over its patience window $W_{\phi_t}$, yielding
\begin{equation}
\chi_t^{\mathrm{stall}}
=
\mathbf{1}\!\left[
\sum_{\tau=t-W_{\phi_t}+1}^{t}
\chi_\tau^{\mathrm{prog}}=0
\right].
\label{eq:stall_indicator}
\end{equation}
We additionally detect phase timeouts using $\chi_t^{\mathrm{timeout}}=\mathbf{1}[n_t^{\mathrm{phase}}>T_{\phi_t}]$, where $n_t^{\mathrm{phase}}$ denotes the elapsed duration of the current phase and $T_{\phi_t}$ its time limit.

The Verifier Agent assigns a diagnosis $d_t$ by evaluating the following conditions in order, giving precedence to target-binding errors.
\begin{equation}
d_t=
\begin{cases}
\text{target-failure},
& \kappa_t<\tau_\kappa
\ \lor\
\mathrm{Inconsistent}(g_t,I_t),\\[1mm]
\text{pick-failure},
& \phi_t\in\Phi_{\mathrm{pick}}
\land
\left(\chi_t^{\mathrm{stall}}=1
\lor\chi_t^{\mathrm{timeout}}=1\right),\\[1mm]
\text{place-failure},
& \phi_t\in\Phi_{\mathrm{place}}
\land
\left(\chi_t^{\mathrm{stall}}=1
\lor\chi_t^{\mathrm{timeout}}=1\right),\\[1mm]
\text{none},
& \text{otherwise}.
\end{cases}
\label{eq:phase_diagnosis}
\end{equation}
Pick failures involve approach, grasping, or lifting, whereas place failures involve transport, placement, release, or post-release stabilization. The resulting diagnosis guides the Correction Agent in selecting the appropriate response.

\subsection{Correction Agent}
\label{sec:execution_correction}

The Correction Agent uses the Verifier Agent's diagnosis to select a failure-specific response. When no intervention is required, the pretrained VLA continues to generate actions from the current observation and grounded prompt. Otherwise, the Correction Agent selects either target rebinding or local execution recovery.

For target failures, the Correction Agent invokes the Grounder Agent to rerun object detection and match candidate instances to the task specification. The updated grounding state $g_t$ is used to regenerate $\tilde{x}_{k_t}$, after which the pretrained VLA resumes control with the corrected target binding. This intervention changes the instance on which the policy acts without modifying the policy itself.

For execution failures, the Correction Agent retains the target lock and invokes the bounded local controller $\pi_{\mathrm{pick}}$ or $\pi_{\mathrm{place}}$. The controller selects a phase-specific waypoint $p_t^*$. It uses a hover point for approach, the grounded target for grasping and lifting, the grounded destination for transport and placement, and a retreat point for release. The translational recovery command is given by
\begin{equation}
u_t^{\mathrm{rec}}
=
\operatorname{clip}
\left(
(p_t^*-p_t^{\mathrm{ee}})\oslash\boldsymbol{s},
-1,1
\right),
\label{eq:recovery_translation}
\end{equation}
where $\oslash$ denotes element-wise division and $\boldsymbol{s}=(s_x,s_y,s_z)$ contains the per-axis displacement scales. The complete recovery action is
\begin{equation}
a_t^{\mathrm{rec}}
=
\left[
\left(u_t^{\mathrm{rec}}\right)^\top,\,
\mathbf{0}_3^\top,\,
a_t^{\mathrm{grip}}
\right]^\top,
\label{eq:recovery_action}
\end{equation}
where $\mathbf{0}_3$ specifies zero rotational commands and $a_t^{\mathrm{grip}}$ denotes the gripper command.
Pick recovery is considered successful when the object moves consistently with the gripper after closure. Place recovery is considered successful when the object is released within the destination region and remains stable. These bounded interventions preserve the pretrained VLA as the default action generator while providing targeted correction for execution failures.

\section{Experiments}
\label{sec:experiments}

\subsection{Experimental Setup}
\label{sec:experimental_setup}

We evaluate FIT-MAC on LIBERO-Object and LIBERO-Spatial~\citep{liu2023libero} following the LIBERO-PRO perturbation protocol~\citep{zhou2026liberopro}. We consider five perturbation types, denoted as Pos, Task, Env, Obj, and Sem, corresponding to changes in object position, task instruction, environment, object identity, and semantic binding, respectively. We compare OpenVLA with and without FIT-MAC using identical task allocations, episode budgets, and random seeds. Published benchmark results for other methods provide additional context.

We use success rate (SR), defined as the fraction of successfully completed episodes, as the primary metric. We additionally report intervention rate (IR) and recovered success rate (RSR). Let $N_{\mathrm{all}}$ denote the total number of evaluated episodes, $N_{\mathrm{int}}$ the number receiving at least one corrective action, and $N_{\mathrm{succ,int}}$ the number of successful episodes among them. We define
\begin{equation}
\mathrm{IR} = \frac{N_{\mathrm{int}}}{N_{\mathrm{all}}},
\qquad
\mathrm{RSR} = \frac{N_{\mathrm{succ,int}}}{N_{\mathrm{int}}}.
\end{equation}
IR measures the proportion of episodes requiring intervention, while RSR measures the success rate among episodes receiving intervention. Together with SR, these metrics characterize intervention usage and task completion.

To assess generality across VLA backbones, we also integrate FIT-MAC with $\pi_0$~\citep{black2025pi0} and $\pi_{0.5}$~\citep{physicalintelligence2025pi05}, keeping all backbone parameters frozen. We conduct module ablations and analyze sensitivity to intervention criteria, recovery timing, and execution horizon. For the controlled OpenVLA comparison, we report results averaged over five random seeds (7, 11, 13, 17, and 23). Complete per-seed results are provided in Appendix~\ref{app:seed_results}.s

\subsection{Main Results}
\label{sec:main_results}

As shown in Table~\ref{tab:libero_pro_comparison}, FIT-MAC achieves state-of-the-art SR in eight of the ten perturbation settings among the compared methods. Its average SR reaches 78.7\%, exceeding OpenVLA by 32.8\%. The largest gains over competing methods occur under position and task perturbations. On LIBERO-Object, FIT-MAC achieves SR of 93.4\% and 87.2\%, outperforming RATs, the strongest competitor in these settings, by 32.4\% and 24.2\%, respectively. On LIBERO-Spatial, FIT-MAC also ranks first under position and task perturbations, achieving SR of 40.2\% and 37.8\%, respectively. Under environment perturbations, it achieves SR of 48.8\% on LIBERO-Object and 97.8\% on LIBERO-Spatial. The corresponding SR under object perturbations is 99.0\% and 97.2\%. Although FIT-MAC does not rank first under semantic perturbations, it outperforms OpenVLA in all ten settings, supporting the effectiveness of selective runtime correction.

In Table~\ref{tab:intervention_recovery}, we report SR, IR, and RSR for each perturbation to characterize intervention frequency and recovery outcomes. On LIBERO-Object, FIT-MAC intervenes frequently under Obj, Pos, Sem, and Task perturbations while maintaining high RSR. On LIBERO-Spatial, OpenVLA achieves zero SR under Pos and Task perturbations, creating substantial demand for corrective intervention. FIT-MAC intervenes in 87.6\% and 90.6\% of episodes, respectively, achieving SR of 40.2\% and 37.8\%. By comparison, OpenVLA already achieves higher SR under Obj, Sem, and Env perturbations, so FIT-MAC intervenes less frequently, with IR of 37.8\%, 42.8\%, and 21.2\%, respectively. RSR exceeds 97\% across these three settings, indicating that most intervened episodes succeed. This pattern is consistent with selective intervention that responds to execution failures while allowing the pretrained policy to retain control when correction is unnecessary.

\begin{figure}[t]
    \centering
    \includegraphics[width=\textwidth]{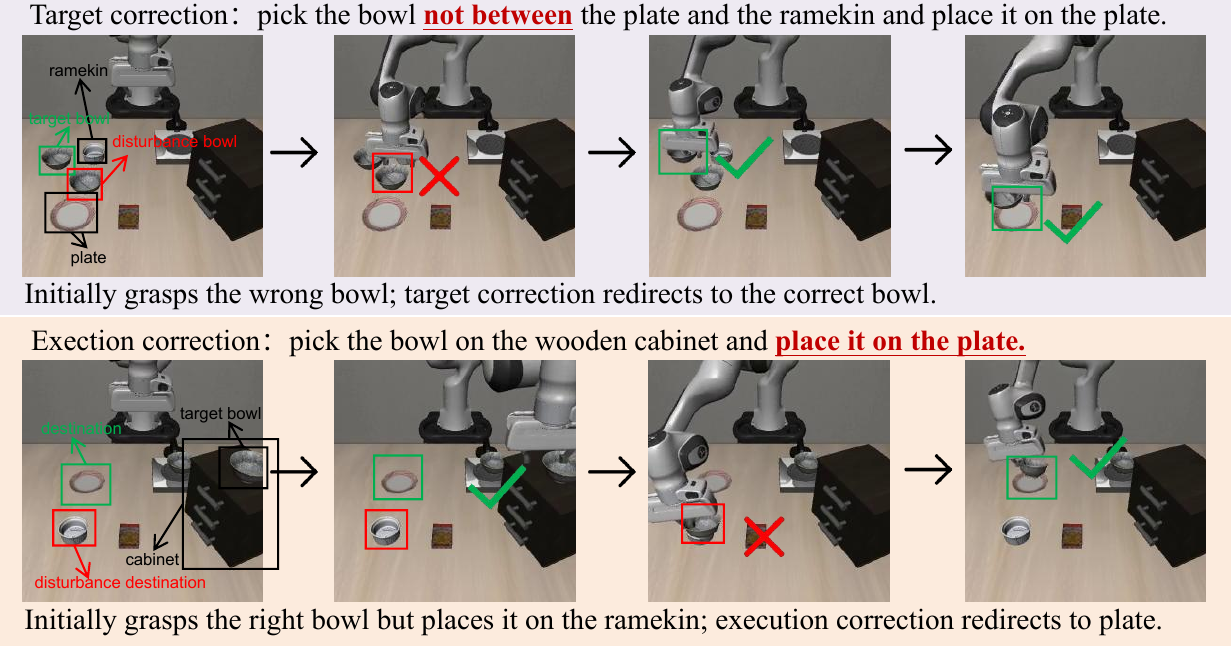}
    \vspace{-15pt}
    \caption{Examples of target correction and execution recovery. The top row shows target rebinding after a grounding failure, the bottom row shows local recovery after a manipulation failure.}
    \vspace{-4pt}
    \label{fig:correction_case}
\end{figure}

In Figure~\ref{fig:correction_case}, we illustrate target correction and execution correction. In the top row, the instruction specifies the bowl that is not between the plate and the ramekin, but the policy initially grasps the bowl between them. FIT-MAC corrects the target binding and redirects manipulation to the intended bowl for placement on the plate. In the bottom row, the policy grasps the correct bowl from the cabinet but places it in the ramekin rather than on the plate. FIT-MAC detects the placement error and invokes local recovery to move the bowl to the specified destination. These examples illustrate how FIT-MAC addresses target-binding errors and local execution failures through distinct corrective mechanisms.

\begin{table}[t]
\centering
\caption{Comparison with state-of-the-art methods on LIBERO-PRO. The overall average is reported for methods evaluated on
all ten perturbation settings.}
\label{tab:libero_pro_comparison}

\small
\setlength{\tabcolsep}{3.5pt}
\renewcommand{\arraystretch}{1.08}

\resizebox{\columnwidth}{!}{%
\begin{tabular}{lccccccccccc}
\toprule

Method
& \multicolumn{5}{c}{LIBERO-Object}
& \multicolumn{5}{c}{LIBERO-Spatial}
& Avg. \\

\cmidrule(lr){2-6}
\cmidrule(lr){7-11}

& Obj.
& Pos.
& Sem.
& Task
& Env.
& Obj.
& Pos.
& Sem.
& Task
& Env.
& \\

\midrule

MolmoAct~\cite{lee2025molmoact}
& 92.0 & 6.0 & 96.0 & 0.0 & --
& 90.0 & 0.0 & 88.0 & 0.0 & --
& -- \\

NORA~\cite{hung2025nora}
& 86.0 & 0.0 & 92.0 & 0.0 & --
& 92.0 & 0.0 & 91.0 & 0.0 & --
& -- \\

X-VLA~\cite{zheng2025xvla}
& 89.0 & 2.0 & 98.0 & 8.0 & --
& 97.0 & 0.0 & \textbf{96.0} & 0.0 & --
& -- \\

AtomVLA~\cite{sun2026atomvla}
& 93.0 & 10.0 & \textbf{99.0} & 0.0 & --
& 95.0 & 16.0 & 95.0 & 1.0 & --
& -- \\

CaP-Agent$_0$~\cite{fu2026capx}
& -- & 22.0 & -- & 18.0 & --
& -- & 12.0 & -- & 14.0 & --
& -- \\

RATs~\cite{zhang2026rats}
& -- & 61.0 & -- & 63.0 & --
& -- & 29.0 & -- & 31.0 & --
& -- \\

OpenVLA~\cite{kim2024openvla}
& 98.0 & 0.0 & 98.0 & 0.0 & 0.0
& 97.0 & 0.0 & 77.0 & 0.0 & 89.0
& 45.9 \\

\midrule

\textbf{FIT-MAC (Ours)}
& \textbf{99.0} & \textbf{93.4} & 98.8
& \textbf{87.2} & \textbf{48.8}
& \textbf{97.2} & \textbf{40.2} & 86.8
& \textbf{37.8} & \textbf{97.8}
& \textbf{78.7} \\

\bottomrule
\end{tabular}%
}
\end{table}

\begin{table}[t]
\centering
\caption{Runtime intervention and recovery behavior of FIT-MAC under different
perturbations. Each cell reports SR/IR/RSR (\%).}
\label{tab:intervention_recovery}

\small
\setlength{\tabcolsep}{4pt}
\renewcommand{\arraystretch}{1.08}

\begin{tabular}{lcccccc}
\toprule
& \multicolumn{3}{c}{LIBERO-Object}
& \multicolumn{3}{c}{LIBERO-Spatial} \\
\cmidrule(lr){2-4}
\cmidrule(lr){5-7}

Perturbation
& SR. (\%)
& IR. (\%)
& RSR. (\%)
& SR. (\%)
& IR. (\%)
& RSR. (\%)  \\
\midrule

Obj.
& 99.0 & 99.8 & 99.3
& 97.2 & 37.8 & 97.9 \\

Pos.
& 93.4 & 99.5 & 93.8
& 40.2 & 87.6 & 41.3 \\

Sem.
& 98.8 & 98.9 & 99.6
& 86.8 & 42.8 & 98.6 \\

Task
& 87.2 & 100.0 & 87.1
& 37.8 & 90.6 & 35.5 \\

Env.
& 48.8 & 50.4 & 97.5
& 97.8 & 21.2 & 99.1 \\

\bottomrule
\end{tabular}
\end{table}

\subsection{Generalization across VLA Backbones}
\label{sec:cross_backbone}
\begin{figure}[t]
    \centering
    \includegraphics[
        width=\textwidth
    ]{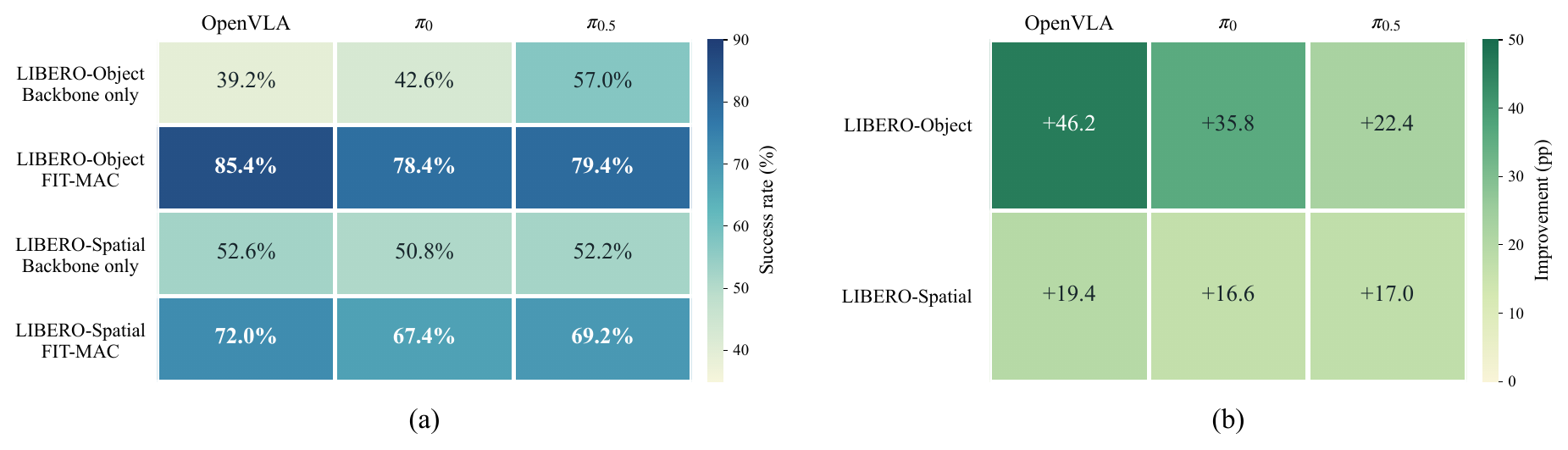}
    \vspace{-25pt}
    \caption{FIT-MAC across VLA backbones on LIBERO-PRO. (a) Success rates of OpenVLA, $\pi_0$, and $\pi_{0.5}$ with and without FIT-MAC, averaged over five perturbation types in each suite. (b) Corresponding absolute gains in success rate.}
    \vspace{-10pt}
    \label{fig:libero_pro_heatmap1}
\end{figure}

In Figure~\ref{fig:libero_pro_heatmap1}(a), we evaluate FIT-MAC with OpenVLA, $\pi_0$, and $\pi_{0.5}$ to assess its generality across VLA backbones. FIT-MAC consistently improves average SR on both suites. On LIBERO-Object, FIT-MAC achieves an average SR of 85.4\%, 78.4\%, and 79.4\% with the three backbones, corresponding to gains of 46.2\%, 35.8\%, and 22.4\%, respectively. On LIBERO-Spatial, the corresponding average SR is 72.0\%, 67.4\%, and 69.2\%, with gains of 19.4\%, 16.6\%, and 17.0\%, respectively. In Figure~\ref{fig:libero_pro_heatmap1}(b), we summarize these gains. These results demonstrate that FIT-MAC improves robustness across different VLA backbones without modifying their parameters.

\subsection{Ablation Study}

\paragraph{Module Ablation.}
In Table~\ref{tab:module_ablation_1}, we assess the contribution of each component through controlled ablations with OpenVLA. We report average SR across five perturbation types for each suite, the mean across both suites (Avg.), and the change relative to full FIT-MAC ($\Delta$SR). Removing any component reduces SR on both suites. Execution correction makes the largest contribution, with its removal reducing average SR from 78.7\% to 54.7\%. The corresponding drops on LIBERO-Object and LIBERO-Spatial are 38.4\% and 9.6\%, respectively.

Hard pick primarily benefits LIBERO-Object, where its removal reduces SR by 35.2\%, compared with 2.0\% on LIBERO-Spatial. Hard place contributes substantially to both suites, with its removal reducing SR by 22.2\% and 8.0\%, respectively. Target correction provides smaller but consistent gains, with its removal reducing SR by 8.2\% and 2.2\%, respectively. These results support the complementary roles of target correction and local execution recovery. Complete per-perturbation results are provided in Appendix~\ref{app:module_ablation}.

\begin{table}[t]
\centering
\caption{Module ablation of FIT-MAC with the OpenVLA backbone on LIBERO-PRO.
}
\label{tab:module_ablation_1}

\small
\setlength{\tabcolsep}{4pt}
\renewcommand{\arraystretch}{1.12}

\begin{tabular}{lccc|cc}
\toprule
\multirow{2}{*}{Configuration}
& \multicolumn{3}{c|}{Success Rate (\%)}
& \multicolumn{2}{c}{$\Delta$SR } \\
\cmidrule(lr){2-4}
\cmidrule(lr){5-6}

& Object
& Spatial
& Avg.
& Object
& Spatial \\
\midrule

FIT-MAC
& \textbf{85.4}
& \textbf{72.0}
& \textbf{78.7}
& 0.0
& 0.0 \\

w/o target correction
& 77.2
& 69.8
& 73.5
& -8.2
& -2.2 \\

w/o execution correction
& 47.0
& 62.4
& 54.7
& -38.4
& -9.6 \\

w/o hard pick
& 50.2
& 70.0
& 60.1
& -35.2
& -2.0 \\

w/o hard place
& 63.2
& 64.0
& 63.6
& -22.2
& -8.0 \\

\bottomrule
\end{tabular}
\end{table}

\paragraph{Hyperparameter Sensitivity Analysis.}
\label{sec:hyperparameter_sensitivity}

In Table~\ref{tab:hyperparameter_ablation}, we assess FIT-MAC's sensitivity to four hyperparameters using OpenVLA. We report average SR across five perturbation types for each suite and their overall mean. Across settings of the first three parameters, SR on LIBERO-Object remains at 85.4\%, while overall average SR varies by at most 0.3\%. 
The execution horizon has a larger effect. Increasing it from $1$ to $2$ raises average SR from 70.0\% to 77.7\%, primarily through gains on LIBERO-Object. A horizon of $3$ further raises average SR to 78.6\%, but reduces SR under position and environment perturbations on LIBERO-Spatial. We therefore retain $2$ for more balanced performance across perturbation types. Complete per-perturbation results are provided in Appendix~\ref{app:hyperparameter_sensitivity}.
\begin{table}[H]
\centering
\caption{Hyperparameter sensitivity of FIT-MAC with the OpenVLA backbone on
LIBERO-PRO. Default values are shown in bold.
}
\vspace{1pt}
\label{tab:hyperparameter_ablation}

\small
\setlength{\tabcolsep}{4pt}
\renewcommand{\arraystretch}{1.15}

\begin{tabular}{lcccc}
\toprule
Parameter
& Values
& Object
& Spatial
& Avg. \\
\midrule

Failure threshold
& 2 / \textbf{3} / 4
& 85.4 / 85.4 / 85.4
& 70.8 / 70.6 / 70.6
& 78.1 / 78.0 / 78.0 \\

Confidence threshold
& 0.10 / \textbf{0.12} / 0.18
& 85.4 / 85.4 / 85.4
& 71.0 / 70.6 / 71.0
& 78.2 / 78.0 / 78.2 \\

Minimum rescue step
& 8 / \textbf{12} / 16
& 85.4 / 85.4 / 85.4
& 71.2 / 70.6 / 71.0
& 78.3 / 78.0 / 78.2 \\

Execution horizon
& 1 / \textbf{2} / 3
& 68.0 / 85.4 / 87.4
& 72.0 / 70.0 / 69.8
& 70.0 / 77.7 / 78.6 \\

\bottomrule
\end{tabular}
\vspace*{-5pt}
\end{table}

\vspace*{-5pt}
\section{Conclusion}
\label{sec:conclusion}

We introduced FIT-MAC, a failure-aware, closed-loop multi-agent control framework that improves the robustness of VLA policies at inference time. FIT-MAC distinguishes target-binding errors from local execution failures and applies targeted corrections without retraining or modifying the pretrained policy. Experiments show consistent robustness gains across VLA backbones and perturbation settings. Intervention and recovery analyses characterize its runtime behavior, while ablations demonstrate the contribution of each component. Future work will focus on reducing intervention overhead and improving semantic and spatial grounding to support reliable long-horizon execution.
\clearpage

\FloatBarrier
\bibliography{iclr2027_conference}
\bibliographystyle{iclr2027_conference}
\clearpage

\appendix
\section{Additional Experimental Results}
\label{app:additional_results}

\subsection{Evaluation Protocol}
\label{app:evaluation_protocol}

We evaluate OpenVLA, $\pi_0$, and $\pi_{0.5}$ with FIT-MAC on LIBERO-Object and LIBERO-Spatial. Each suite includes five perturbation conditions, namely Task, Pos, Obj, Sem, and Env. For each method and perturbation setting, we report success rate (SR), intervention rate (IR), and recovered success rate (RSR). Under each evaluation condition, we run 50 episodes per task for the main experiments and 10 episodes per task for the ablations. For multi-seed experiments, we use five random seeds, $\{7, 11, 13, 17, 23\}$, with 10 episodes per task per seed under each condition.

\subsection{Complete Cross-Backbone Results}
\label{app:cross_backbone_results}

In Table~\ref{tab:cross_backbone_full}, we report complete SR, IR, and RSR results for OpenVLA, $\pi_0$, and $\pi_{0.5}$ across both suites and all five perturbation settings. In Table~\ref{tab:cross_backbone}, we summarize average SR and absolute gains over the corresponding backbone-only baselines. On LIBERO-Object, FIT-MAC increases average SR for OpenVLA, $\pi_0$, and $\pi_{0.5}$ from 39.2\%, 42.6\%, and 57.0\% to 85.4\%, 78.4\%, and 79.4\%, respectively. The corresponding absolute gains are 46.2\%, 35.8\%, and 22.4\%. On LIBERO-Spatial, FIT-MAC raises average SR from 52.6\%, 50.8\%, and 52.2\% to 72.0\%, 67.4\%, and 69.2\%, yielding absolute gains of 19.4\%, 16.6\%, and 17.0\%, respectively. Although the magnitude of improvement varies across backbones and suites, FIT-MAC consistently improves average SR for all three backbones.

\begin{table}[H]
\centering
\scriptsize
\setlength{\tabcolsep}{3pt}
\renewcommand{\arraystretch}{1.1}
\caption{Complete cross-backbone results under five perturbation conditions. Each cell reports SR/IR/RSR, all expressed as percentages (\%).}
\label{tab:cross_backbone_full}
\begin{tabular}{llcccccc}
\toprule
\textbf{Benchmark}
& \textbf{Condition}
& \textbf{OpenVLA}
& \textbf{OpenVLA + FIT-MAC}
& \textbf{$\pi$0}
& \textbf{$\pi$0 + FIT-MAC}
& \textbf{$\pi$0.5}
& \textbf{$\pi$0.5 + FIT-MAC} \\
\midrule

\multirow{5}{*}{LIBERO-Object}

& Task
& 0.0/--/--
& 87.3/100.0/87.1
& 0.0/--/--
& 84.0/98.0/85.7
& 1.0/--/--
& 80.0/100.0/80.0 \\

& Pos
& 0.0/--/--
& 93.1/99.5/93.8
& 0.0/--/--
& 67.0/100.0/67.0
& 17.0/--/--
& 67.0/99.0/67.7 \\

& Obj
& 98.0/--/--
& 99.0/99.8/99.3
& 94.0/--/--
& 89.0/100.0/89.0
& 98.0/--/--
& 92.0/100.0/92.0 \\

& Sem
& 98.0/--/--
& 98.8/98.9/99.6
& 90.0/--/--
& 87.0/97.0/89.7
& 96.0/--/--
& 89.0/97.0/91.8 \\

& Env
& 0.0/--/--
& 48.8/50.4/97.5
& 29.0/--/--
& 65.0/78.0/82.1
& 73.0/--/--
& 69.0/79.0/86.1 \\

\midrule

\multirow{5}{*}{LIBERO-Spatial}

& Task
& 0.0/--/--
& 37.8/90.6/35.5
& 0.0/--/--
& 27.0/100.0/27.0
& 1.0/--/--
& 25.0/100.0/25.0 \\

& Pos
& 0.0/--/--
& 40.2/87.6/41.3
& 0.0/--/--
& 36.0/98.0/34.7
& 20.0/--/--
& 37.0/98.0/37.8 \\

& Obj
& 97.0/--/--
& 97.2/37.8/97.9
& 97.0/--/--
& 97.0/52.0/98.1
& 97.0/--/--
& 97.0/52.0/100.0 \\

& Sem
& 77.0/--/--
& 86.8/42.8/98.6
& 97.0/--/--
& 83.0/60.0/81.7
& 97.0/--/--
& 93.0/63.0/88.9 \\

& Env
& 89.0/--/--
& 97.8/21.2/99.1
& 60.0/--/--
& 94.0/42.0/95.2
& 46.0/--/--
& 94.0/41.0/97.6 \\

\bottomrule
\end{tabular}
\end{table}

\begin{table}[H]
\centering
\caption{Average success rates (\%) of FIT-MAC with different VLA backbones.
Results are averaged over the five perturbation types within each benchmark.
The $\Delta$ rows report absolute gains 
over the corresponding backbone-only baselines.}
\label{tab:cross_backbone}
\small
\setlength{\tabcolsep}{5pt}
\renewcommand{\arraystretch}{1.08}

\begin{tabular}{llccc}
\toprule
Benchmark
& Method
& OpenVLA
& $\pi_0$
& $\pi_{0.5}$ \\
\midrule

\multirow{3}{*}{LIBERO-Object}
& Backbone only
& 39.2
& 42.6
& 57.0 \\

& FIT-MAC
& \textbf{85.4}
& \textbf{78.4}
& \textbf{79.4} \\

& $\Delta$
& +46.2
& +35.8
& +22.4 \\

\midrule

\multirow{3}{*}{LIBERO-Spatial}
& Backbone only
& 52.6
& 50.8
& 52.2 \\

& FIT-MAC
& \textbf{72.0}
& \textbf{67.4}
& \textbf{69.2} \\

& $\Delta$ 
& +19.4
& +16.6
& +17.0 \\

\bottomrule
\end{tabular}
\end{table}

\subsection{Multi-Seed Results}
\label{app:seed_results}

We evaluate FIT-MAC with OpenVLA on LIBERO-Object and LIBERO-Spatial using five random seeds, $\{7, 11, 13, 17, 23\}$. For each seed, we run 10 episodes per task under each perturbation condition, yielding 100 episodes per condition per suite.
For each suite, perturbation condition, and metric, we compute the mean and population standard deviation across seeds. Let $x_i$ denote the result for seed $i$, expressed as a percentage, and let $N=5$. We define
\begin{equation}
\bar{x}
=
\frac{1}{N}\sum_{i=1}^{N} x_i,
\qquad
\sigma_x
=
\sqrt{
\frac{1}{N}
\sum_{i=1}^{N}
\left(x_i-\bar{x}\right)^2
}.
\label{eq:seed_statistics}
\end{equation}
For suite-level summaries, we first average SR across the five perturbation conditions for each seed, then compute these statistics across seeds.

\paragraph{Summary.}
In Tables~\ref{tab:seed_object} and~\ref{tab:seed_spatial}, we report results across the five seeds. FIT-MAC with OpenVLA achieves average SR of 85.4\% on LIBERO-Object and 71.7\% on LIBERO-Spatial, with population standard deviations of 0.1\% and 0.9\%, respectively. The largest condition-specific variation in SR occurs under position perturbations on LIBERO-Spatial, with a mean of 38.0\% and a standard deviation of 2.8\%. In contrast, SR under object perturbations on LIBERO-Object remains at 100.0\% across all seeds. These results indicate limited variability in SR across the tested seeds under the current evaluation setup.

\begin{table}[t]
\centering
\caption{
Multi-seed results of FIT-MAC on LIBERO-Object with OpenVLA.
Each cell reports SR/IR/RSR (\%).
}
\label{tab:seed_object}

\scriptsize
\setlength{\tabcolsep}{3.5pt}
\renewcommand{\arraystretch}{1.08}

\begin{tabular}{llccccc}
\toprule
\textbf{Method}
& \textbf{Seed}
& \textbf{Obj.}
& \textbf{Pos.}
& \textbf{Sem.}
& \textbf{Task}
& \textbf{Env.} \\
\midrule

\multirow{5}{*}{OpenVLA + FIT-MAC}
& 7
& 100.0/100.0/100.0
& 93.0/100.0/93.0
& 98.0/98.0/100.0
& 87.0/100.0/87.0
& 49.0/49.0/100.0 \\

& 11
& 100.0/100.0/100.0
& 94.0/99.0/94.9
& 97.0/97.0/100.0
& 86.0/100.0/86.0
& 49.0/49.0/100.0 \\

& 13
& 100.0/100.0/100.0
& 93.0/100.0/93.0
& 98.0/98.0/100.0
& 87.0/100.0/87.0
& 49.0/49.0/100.0 \\

& 17
& 100.0/100.0/100.0
& 94.0/99.0/94.9
& 98.0/98.0/100.0
& 86.0/100.0/86.0
& 49.0/49.0/100.0 \\

& 23
& 100.0/100.0/100.0
& 94.0/100.0/94.0
& 98.0/98.0/100.0
& 87.0/100.0/87.0
& 48.0/48.0/100.0 \\

\bottomrule
\end{tabular}
\end{table}

\begin{table}[t]
\centering
\caption{
Multi-seed results of FIT-MAC on LIBERO-Spatial with OpenVLA.
Each cell reports SR/IR/RSR (\%).
}
\label{tab:seed_spatial}

\scriptsize
\setlength{\tabcolsep}{3.5pt}
\renewcommand{\arraystretch}{1.08}

\begin{tabular}{llccccc}
\toprule
\textbf{Method}
& \textbf{Seed}
& \textbf{Obj.}
& \textbf{Pos.}
& \textbf{Sem.}
& \textbf{Task}
& \textbf{Env.} \\
\midrule

\multirow{5}{*}{OpenVLA + FIT-MAC}
& 7
& 97.0/36.0/100.0
& 38.0/89.0/40.4
& 86.0/42.0/95.2
& 40.0/96.0/39.6
& 99.0/28.0/100.0 \\

& 11
& 96.0/39.0/100.0
& 34.0/87.0/36.8
& 88.0/41.0/100.0
& 40.0/95.0/38.9
& 96.0/24.0/100.0 \\

& 13
& 98.0/37.0/97.3
& 41.0/90.0/40.0
& 88.0/38.0/97.4
& 35.0/95.0/32.6
& 97.0/19.0/100.0 \\

& 17
& 100.0/37.0/100.0
& 41.0/91.0/40.7
& 88.0/44.0/95.5
& 38.0/94.0/38.3
& 99.0/24.0/100.0 \\

& 23
& 98.0/32.0/100.0
& 36.0/91.0/38.5
& 85.0/45.0/95.6
& 38.0/90.0/35.6
& 97.0/27.0/96.3 \\

\bottomrule
\end{tabular}
\end{table}

\subsection{Complete Module Ablation}
\label{app:module_ablation}

We conduct controlled module ablations with OpenVLA and $\pi_{0.5}$ to assess the contribution of each FIT-MAC component. We compare full FIT-MAC, which includes both correctors and both constraints, with four variants that individually remove the target corrector, execution corrector, hard pick, or hard place. Each variant keeps the backbone, prompts, evaluation seeds, perturbation settings, and remaining components unchanged.

In Table~\ref{tab:module_ablation_object} and Table~\ref{tab:module_ablation_spatial}, we report results on LIBERO-Object and LIBERO-Spatial, respectively. Each cell reports success rate (SR), intervention rate (IR), and recovered success rate (RSR). For backbone-only baselines, IR and RSR are marked as ``--''. For FIT-MAC variants without interventions, IR is reported as 0.0\% and RSR as ``--''.

Removing the target corrector reduces SR under position perturbations for both backbones on both suites. On LIBERO-Object, SR decreases from 94.0\% to 70.0\% for OpenVLA and from 67.0\% to 59.0\% for $\pi_{0.5}$. On LIBERO-Spatial, the corresponding decreases are from 41.0\% to 25.0\% and from 37.0\% to 30.0\%. Removing the execution corrector causes larger drops under position perturbations on LIBERO-Object, reducing SR to 4.0\% for OpenVLA and 11.0\% for $\pi_{0.5}$. For OpenVLA, it also reduces SR under task perturbations from 86.0\% to 0.0\%.

Hard pick provides substantial gains under task perturbations on LIBERO-Object, where its removal reduces SR from 86.0\% to 0.0\% for OpenVLA and from 80.0\% to 13.0\% for $\pi_{0.5}$. Hard place improves performance under position perturbations on LIBERO-Spatial, where its removal reduces SR from 41.0\% to 5.0\% and from 37.0\% to 14.0\%, respectively. However, some ablated variants outperform full FIT-MAC in individual settings. For $\pi_{0.5}$ on LIBERO-Spatial, removing hard pick increases SR from 25.0\% to 45.0\% under task perturbations and from 37.0\% to 55.0\% under position perturbations. These results identify settings in which each component contributes while showing that its benefits depend on the backbone and perturbation.

\begin{table}[t]
\centering
\scriptsize
\setlength{\tabcolsep}{2.5pt}
\renewcommand{\arraystretch}{1.08}

\caption{Module ablation results under different perturbation conditions on LIBERO-Object. Each cell reports SR/IR/RSR (\%).}
\label{tab:module_ablation_object}

\begin{tabular}{llccccc}
\toprule
\textbf{Backbone}
& \textbf{Configuration}
& \textbf{Task}
& \textbf{Pos.}
& \textbf{Obj.}
& \textbf{Sem.}
& \textbf{Env.} \\
\midrule

\multirow{6}{*}{OpenVLA}
& Baseline
& 0.0/--/--
& 0.0/--/--
& 98.0/--/--
& 98.0/--/--
& 0.0/--/-- \\

& Full FIT-MAC
& 86.0/100.0/86.0
& 94.0/99.0/94.9
& 100.0/100.0/100.0
& 98.0/98.0/100.0
& 49.0/49.0/100.0 \\

& w/o target\_corrector
& 69.0/80.0/86.3
& 70.0/75.0/93.3
& 100.0/100.0/100.0
& 98.0/98.0/100.0
& 49.0/49.0/100.0 \\

& w/o execution\_corrector
& 0.0/0.0/--
& 4.0/0.0/--
& 100.0/0.0/--
& 97.0/0.0/--
& 34.0/0.0/-- \\

& w/o hard\_pick
& 0.0/0.0/--
& 4.0/4.0/100.0
& 100.0/100.0/100.0
& 98.0/98.0/100.0
& 49.0/49.0/100.0 \\

& w/o hard\_place
& 25.0/100.0/25.0
& 60.0/96.0/58.3
& 100.0/0.0/--
& 97.0/0.0/--
& 34.0/0.0/-- \\

\midrule

\multirow{6}{*}{$\pi$0.5}
& Baseline
& 1.0/--/--
& 17.0/--/--
& 98.0/--/--
& 96.0/--/--
& 73.0/--/-- \\

& Full FIT-MAC
& 80.0/100.0/80.0
& 67.0/99.0/67.7
& 92.0/100.0/92.0
& 89.0/97.0/91.8
& 69.0/79.0/86.1 \\

& w/o target\_corrector
& 71.0/83.0/85.5
& 59.0/94.0/62.8
& 90.0/100.0/90.0
& 87.0/100.0/87.0
& 67.0/79.0/84.8 \\

& w/o execution\_corrector
& 8.0/0.0/--
& 11.0/0.0/--
& 92.0/0.0/--
& 92.0/0.0/--
& 53.0/0.0/-- \\

& w/o hard\_pick
& 13.0/13.0/100.0
& 11.0/13.0/84.6
& 96.0/100.0/96.0
& 91.0/98.0/92.9
& 56.0/75.0/74.7 \\

& w/o hard\_place
& 36.0/89.0/31.5
& 45.0/100.0/45.0
& 97.0/0.0/--
& 89.0/0.0/--
& 55.0/0.0/-- \\

\bottomrule
\end{tabular}
\end{table}

\begin{table}[t]
\centering
\scriptsize
\setlength{\tabcolsep}{2.5pt}
\renewcommand{\arraystretch}{1.08}

\caption{Module ablation results under different perturbation conditions on LIBERO-Spatial. Each cell reports SR/IR/RSR (\%).}
\label{tab:module_ablation_spatial}

\begin{tabular}{llccccc}
\toprule
\textbf{Backbone}
& \textbf{Configuration}
& \textbf{Task}
& \textbf{Pos.}
& \textbf{Obj.}
& \textbf{Sem.}
& \textbf{Env.} \\
\midrule

\multirow{6}{*}{OpenVLA}
& Baseline
& 0.0/--/--
& 0.0/--/--
& 97.0/--/--
& 77.0/--/--
& 89.0/--/-- \\

& Full FIT-MAC
& 38.0/94.0/38.3
& 41.0/88.0/40.9
& 97.0/34.0/100.0
& 85.0/44.0/97.7
& 99.0/24.0/100.0 \\

& w/o target\_corrector
& 41.0/79.0/36.7
& 25.0/78.0/25.6
& 98.0/26.0/100.0
& 86.0/33.0/100.0
& 99.0/12.0/100.0 \\

& w/o execution\_corrector
& 28.0/0.0/--
& 5.0/0.0/--
& 95.0/0.0/--
& 86.0/0.0/--
& 98.0/0.0/-- \\

& w/o hard\_pick
& 28.0/40.0/35.0
& 41.0/37.0/97.3
& 97.0/34.0/100.0
& 85.0/44.0/97.7
& 99.0/24.0/100.0 \\

& w/o hard\_place
& 36.0/80.0/26.3
& 5.0/70.0/0.0
& 95.0/0.0/--
& 86.0/0.0/--
& 98.0/0.0/-- \\

\midrule

\multirow{6}{*}{$\pi$0.5}
& Baseline
& 1.0/--/--
& 20.0/--/--
& 97.0/--/--
& 97.0/--/--
& 46.0/--/-- \\

& Full FIT-MAC
& 25.0/100.0/25.0
& 37.0/98.0/37.8
& 97.0/52.0/100.0
& 93.0/63.0/88.9
& 94.0/41.0/97.6 \\

& w/o target\_corrector
& 27.0/100.0/27.0
& 30.0/97.0/28.9
& 95.0/52.0/98.1
& 90.0/64.0/84.4
& 97.0/42.0/100.0 \\

& w/o execution\_corrector
& 47.0/0.0/--
& 35.0/0.0/--
& 96.0/0.0/--
& 95.0/0.0/--
& 96.0/0.0/-- \\

& w/o hard\_pick
& 45.0/60.0/56.7
& 55.0/38.0/86.8
& 95.0/52.0/96.2
& 90.0/63.0/84.1
& 98.0/42.0/100.0 \\

& w/o hard\_place
& 35.0/100.0/35.0
& 14.0/94.0/10.6
& 97.0/0.0/--
& 96.0/0.0/--
& 96.0/0.0/-- \\

\bottomrule
\end{tabular}
\end{table}

\subsection{Complete Hyperparameter Ablation}
\label{app:hyperparameter_sensitivity}

We assess FIT-MAC's sensitivity to four control parameters governing the failure-streak requirement, target-confidence requirement, recovery timing, and execution horizon. We vary one parameter at a time while keeping the backbone, benchmark, perturbation type, random seed, and all other settings fixed. Each cell reports success rate (SR), intervention rate (IR), and recovered success rate (RSR) for the backbone specified in the corresponding table. These analyses examine how intervention gating, recovery timing, and policy query frequency affect task completion and intervention behavior.

\subsubsection{Intervention-Gate Sensitivity}
\label{app:hyperparameter_gate}

The parameter \text{multi-agent-adaptive-intervention-failures} specifies the minimum number of consecutive actionable failures of the same type required by the adaptive intervention gate. A failure is counted only when the current attempt is flagged for possible termination and diagnosed with a target or execution failure. The count resets when the diagnosed failure type changes.
The gate also requires a sufficient sequence of high-confidence target observations. Thus, this parameter controls only the failure-streak requirement. Intervention additionally depends on other gating conditions and activation signals from strong-evidence and recovery paths.

In Table~\ref{tab:hyper_gate_object} and Table~\ref{tab:hyper_gate_spatial}, we report results for thresholds of $\{2, 3, 4\}$ on LIBERO-Object and LIBERO-Spatial, respectively. Lower thresholds allow earlier responses to repeated failures, whereas higher thresholds require more persistent failure evidence. For OpenVLA on LIBERO-Object, SR remains unchanged across the tested thresholds under all five perturbation conditions. On LIBERO-Spatial, increasing the threshold from $2$ to $3$ reduces SR under task perturbations from 36.0\% to 31.0\%, while increasing SR under position and environment perturbations from 39.0\% to 41.0\% and from 97.0\% to 99.0\%, respectively. Results for $\pi_{0.5}$ show greater variation across thresholds. On LIBERO-Object, a threshold of $3$ yields the highest SR under position perturbations at 78.0\%, whereas a threshold of $4$ performs best under environment perturbations at 84.0\%. These results indicate that no single threshold consistently performs best across backbones and perturbation conditions.

\begin{table}[t]
\centering
\scriptsize
\setlength{\tabcolsep}{2.5pt}
\renewcommand{\arraystretch}{1.08}

\caption{Sensitivity to the intervention-gate threshold on LIBERO-Object.
Each cell reports SR/IR/RSR (\%).}
\label{tab:hyper_gate_object}

\begin{tabular}{llcccccc}
\toprule
\textbf{Backbone}
& \textbf{Setting}
& \textbf{Value}
& \textbf{Task}
& \textbf{Pos.}
& \textbf{Obj.}
& \textbf{Sem.}
& \textbf{Env.} \\
\midrule

\multirow{4}{*}{OpenVLA}
& FIT-MAC
& --
& 86.0/100.0/86.0
& 94.0/99.0/94.9
& 100.0/100.0/100.0
& 98.0/98.0/100.0
& 49.0/49.0/100.0 \\

& failures
& 2
& 86.0/100.0/86.0
& 94.0/99.0/94.9
& 100.0/100.0/100.0
& 98.0/98.0/100.0
& 49.0/49.0/100.0 \\

& failures
& 3
& 86.0/100.0/86.0
& 94.0/100.0/94.0
& 100.0/100.0/100.0
& 98.0/98.0/100.0
& 49.0/49.0/100.0 \\

& failures
& 4
& 86.0/100.0/86.0
& 94.0/100.0/94.0
& 100.0/100.0/100.0
& 98.0/98.0/100.0
& 49.0/49.0/100.0 \\

\midrule

\multirow{4}{*}{$\pi$0.5}
& FIT-MAC
& --
& 80.0/100.0/80.0
& 67.0/99.0/67.7
& 92.0/100.0/92.0
& 89.0/97.0/91.8
& 69.0/79.0/86.1 \\

& failures
& 2
& 66.0/84.0/78.6
& 71.0/95.0/74.7
& 91.0/99.0/91.9
& 85.0/99.0/85.9
& 81.0/84.0/96.4 \\

& failures
& 3
& 69.0/85.0/81.2
& 78.0/94.0/83.0
& 93.0/99.0/93.9
& 92.0/98.0/93.9
& 79.0/84.0/92.9 \\

& failures
& 4
& 68.0/84.0/81.0
& 73.0/97.0/75.3
& 95.0/99.0/96.0
& 90.0/99.0/90.9
& 84.0/87.0/95.4 \\

\bottomrule
\end{tabular}
\end{table}

\begin{table}[t]
\centering
\scriptsize
\setlength{\tabcolsep}{2.5pt}
\renewcommand{\arraystretch}{1.08}

\caption{Sensitivity to the intervention-gate threshold on LIBERO-Spatial.
Each cell reports SR/IR/RSR (\%).}
\label{tab:hyper_gate_spatial}

\begin{tabular}{llcccccc}
\toprule
\textbf{Backbone}
& \textbf{Setting}
& \textbf{Value}
& \textbf{Task}
& \textbf{Pos.}
& \textbf{Obj.}
& \textbf{Sem.}
& \textbf{Env.} \\
\midrule

\multirow{4}{*}{OpenVLA}
& FIT-MAC
& --
& 38.0/94.0/38.3
& 41.0/88.0/40.9
& 97.0/34.0/100.0
& 85.0/44.0/97.7
& 99.0/24.0/100.0 \\

& failures
& 2
& 36.0/95.0/35.8
& 39.0/89.0/41.6
& 97.0/31.0/100.0
& 85.0/46.0/97.8
& 97.0/27.0/100.0 \\

& failures
& 3
& 31.0/94.0/30.9
& 41.0/88.0/40.9
& 97.0/34.0/100.0
& 85.0/44.0/97.7
& 99.0/24.0/100.0 \\

& failures
& 4
& 31.0/94.0/30.9
& 41.0/88.0/40.9
& 97.0/34.0/100.0
& 85.0/44.0/97.7
& 99.0/24.0/100.0 \\

\midrule

\multirow{4}{*}{$\pi$0.5}
& FIT-MAC
& --
& 25.0/100.0/25.0
& 37.0/98.0/37.8
& 97.0/52.0/100.0
& 93.0/63.0/88.9
& 94.0/41.0/97.6 \\

& failures
& 2
& 39.0/100.0/39.0
& 51.0/97.0/49.5
& 95.0/52.0/94.2
& 94.0/62.0/91.9
& 95.0/42.0/95.2 \\

& failures
& 3
& 39.0/100.0/39.0
& 41.0/89.0/33.7
& 94.0/52.0/94.2
& 93.0/63.0/93.7
& 96.0/42.0/100.0 \\

& failures
& 4
& 38.0/100.0/38.0
& 42.0/89.0/34.8
& 98.0/52.0/100.0
& 92.0/63.0/90.5
& 94.0/42.0/97.6 \\

\bottomrule
\end{tabular}
\end{table}

\subsubsection{Target-Confidence Gate Sensitivity}
\label{app:hyperparameter_confidence}

The parameter \text{multi-agent-adaptive-intervention-confidence} specifies the minimum target confidence required by the adaptive intervention gate. Confidence may be derived from target probes, shadow grounding, live grounding, and monitoring signals. This threshold determines whether target evidence is sufficiently reliable to inform intervention decisions and is distinct from the confidence thresholds used by individual grounding or VLM components.
In Table~\ref{tab:hyper_confidence_object} and Table~\ref{tab:hyper_confidence_spatial}, we report results for thresholds of $\{0.10, 0.12, 0.18\}$ on LIBERO-Object and LIBERO-Spatial, respectively. Lower thresholds admit target evidence with lower confidence, whereas higher thresholds impose stricter confidence requirements.

For OpenVLA on LIBERO-Object, SR remains unchanged across the tested thresholds under all five perturbation conditions. On LIBERO-Spatial, SR varies from 31.0\% to 37.0\% under task perturbations, while remaining at 85.0\% under semantic perturbations. For $\pi_{0.5}$ on LIBERO-Object, increasing the threshold from $0.10$ to $0.18$ reduces SR under semantic perturbations from 88.0\% to 79.0\% and under environment perturbations from 83.0\% to 70.0\%. In contrast, SR under position perturbations remains at 71.0\% across all three thresholds. On LIBERO-Spatial, SR varies by at most 3.0\% within each perturbation condition for $\pi_{0.5}$. These results indicate that sensitivity depends on the backbone and perturbation condition, and stricter confidence requirements do not consistently improve SR.

\begin{table}[t]
\centering
\scriptsize
\setlength{\tabcolsep}{2.5pt}
\renewcommand{\arraystretch}{1.08}
\caption{Sensitivity to the confidence threshold on LIBERO-Object.
Each cell reports SR/IR/RSR.}
\label{tab:hyper_confidence_object}

\begin{tabular}{llcccccc}
\toprule
\textbf{Backbone} & \textbf{Setting} & \textbf{Value}
& \textbf{Task} & \textbf{Pos.} & \textbf{Obj.} & \textbf{Sem.} & \textbf{Env.} \\
\midrule

\multirow{4}{*}{OpenVLA}
& FIT-MAC & -- & 86.0/100.0/86.0 & 94.0/99.0/94.9 & 100.0/100.0/100.0 & 98.0/98.0/100.0 & 49.0/49.0/100.0 \\
& confidence & 0.10 & 86.0/100.0/86.0 & 94.0/99.0/94.9 & 100.0/100.0/100.0 & 98.0/98.0/100.0 & 49.0/49.0/100.0 \\
& confidence & 0.12 & 86.0/100.0/86.0 & 94.0/100.0/94.0 & 100.0/100.0/100.0 & 98.0/98.0/100.0 & 49.0/49.0/100.0 \\
& confidence & 0.18 & 86.0/100.0/86.0 & 94.0/99.0/94.9 & 100.0/100.0/100.0 & 98.0/98.0/100.0 & 49.0/49.0/100.0 \\

\midrule

\multirow{4}{*}{$\pi$0.5}
& FIT-MAC & -- & 80.0/100.0/80.0 & 67.0/99.0/67.7 & 92.0/100.0/92.0 & 89.0/97.0/91.8 & 69.0/79.0/86.1 \\
& confidence & 0.10 & 69.0/85.0/81.2 & 71.0/95.0/74.7 & 91.0/100.0/91.0 & 88.0/100.0/88.0 & 83.0/86.0/95.3 \\
& confidence & 0.12 & 70.0/85.0/82.4 & 71.0/94.0/75.5 & 89.0/99.0/89.9 & 84.0/99.0/84.8 & 79.0/85.0/92.9 \\
& confidence & 0.18 & 68.0/85.0/80.0 & 71.0/94.0/75.5 & 91.0/100.0/91.0 & 79.0/99.0/79.8 & 70.0/91.0/76.9 \\

\bottomrule
\end{tabular}
\end{table}

\begin{table}[t]
\centering
\scriptsize
\setlength{\tabcolsep}{2.5pt}
\renewcommand{\arraystretch}{1.08}
\caption{Sensitivity to the confidence threshold on LIBERO-Spatial.
Each cell reports SR/IR/RSR.}
\label{tab:hyper_confidence_spatial}

\begin{tabular}{llcccccc}
\toprule
\textbf{Backbone} & \textbf{Setting} & \textbf{Value}
& \textbf{Task} & \textbf{Pos.} & \textbf{Obj.} & \textbf{Sem.} & \textbf{Env.} \\
\midrule

\multirow{4}{*}{OpenVLA}
& FIT-MAC & -- & 38.0/94.0/38.3 & 41.0/88.0/40.9 & 97.0/34.0/100.0 & 85.0/44.0/97.7 & 99.0/24.0/100.0 \\
& confidence & 0.10 & 37.0/95.0/36.8 & 39.0/90.0/41.1 & 97.0/31.0/100.0 & 85.0/46.0/97.8 & 97.0/27.0/100.0 \\
& confidence & 0.12 & 31.0/94.0/30.9 & 41.0/88.0/40.9 & 97.0/33.0/100.0 & 85.0/44.0/97.7 & 99.0/24.0/100.0 \\
& confidence & 0.18 & 36.0/95.0/35.8 & 39.0/90.0/41.1 & 98.0/32.0/100.0 & 85.0/46.0/97.8 & 97.0/27.0/100.0 \\
\midrule

\multirow{4}{*}{$\pi$0.5}
& FIT-MAC & -- & 25.0/100.0/25.0 & 37.0/98.0/37.8 & 97.0/52.0/100.0 & 93.0/63.0/88.9 & 94.0/41.0/97.6 \\
& confidence & 0.10 & 41.0/100.0/41.0 & 44.0/89.0/37.1 & 96.0/52.0/98.1 & 93.0/64.0/90.6 & 96.0/42.0/97.6 \\
& confidence & 0.12 & 39.0/100.0/39.0 & 41.0/92.0/35.9 & 98.0/52.0/98.1 & 93.0/63.0/88.9 & 97.0/42.0/97.6 \\
& confidence & 0.18 & 40.0/100.0/40.0 & 43.0/89.0/36.0 & 97.0/51.0/98.0 & 94.0/62.0/93.5 & 96.0/42.0/97.6 \\

\bottomrule
\end{tabular}
\end{table}

\subsubsection{Rescue-Timing Sensitivity}
\label{app:hyperparameter_rescue}

The parameter \text{multi-agent-adaptive-execution-rescue-min-step} specifies the earliest execution step at which time-based evidence can inform execution recovery. This delay helps prevent early transient deviations from being mistaken for execution failures. Recovery additionally requires phase-specific evidence, such as persistent wrong-target behavior, stagnation during later stages of picking or placement, or lack of manipulation progress. Reaching the threshold alone does not trigger recovery.

In Table~\ref{tab:hyper_rescue_object} and Table~\ref{tab:hyper_rescue_spatial}, we report results for thresholds of $\{8, 12, 16\}$ on LIBERO-Object and LIBERO-Spatial, respectively. Lower thresholds allow earlier consideration of time-based recovery evidence, whereas higher thresholds give the current execution more time to make progress.

For OpenVLA on LIBERO-Object, SR remains unchanged across the tested thresholds under all five perturbation conditions. On LIBERO-Spatial, a threshold of $12$ yields SR of 41.0\% under position perturbations and 99.0\% under environment perturbations, compared with 39.0\% and 97.0\%, respectively, at both $8$ and $16$. However, SR under task perturbations decreases to 31.0\% at $12$, compared with 37.0\% at $8$ and 36.0\% at $16$. For $\pi_{0.5}$ on LIBERO-Object, SR under task perturbations is 68.0\%, 63.0\%, and 68.0\% at thresholds of $8$, $12$, and $16$, respectively. On LIBERO-Spatial, increasing the threshold from $8$ to $16$ raises SR under task perturbations from 36.0\% to 41.0\% and under position perturbations from 41.0\% to 44.0\%. These results show that the effect of recovery timing varies across backbones and perturbation conditions, with no threshold consistently yielding the highest SR.

\begin{table}[t]
\centering
\scriptsize
\setlength{\tabcolsep}{2.5pt}
\renewcommand{\arraystretch}{1.08}
\caption{Sensitivity to the minimum rescue step on LIBERO-Object.
Each cell reports SR/IR/RSR.}
\label{tab:hyper_rescue_object}

\begin{tabular}{llcccccc}
\toprule
\textbf{Backbone} & \textbf{Setting} & \textbf{Value}
& \textbf{Task} & \textbf{Pos.} & \textbf{Obj.} & \textbf{Sem.} & \textbf{Env.} \\
\midrule

\multirow{4}{*}{OpenVLA}
& FIT-MAC & -- & 86.0/100.0/86.0 & 94.0/99.0/94.9 & 100.0/100.0/100.0 & 98.0/98.0/100.0 & 49.0/49.0/100.0 \\
& rescue\_min\_step & 8 & 86.0/100.0/86.0 & 94.0/99.0/94.9 & 100.0/100.0/100.0 & 98.0/98.0/100.0 & 49.0/49.0/100.0 \\
& rescue\_min\_step & 12 & 86.0/100.0/86.0 & 94.0/100.0/94.0 & 100.0/100.0/100.0 & 98.0/98.0/100.0 & 49.0/49.0/100.0 \\
& rescue\_min\_step & 16 & 86.0/100.0/86.0 & 94.0/99.0/94.9 & 100.0/100.0/100.0 & 98.0/98.0/100.0 & 49.0/49.0/100.0 \\

\midrule

\multirow{4}{*}{$\pi$0.5}
& FIT-MAC & -- & 80.0/100.0/80.0 & 67.0/99.0/67.7 & 92.0/100.0/92.0 & 89.0/97.0/91.8 & 69.0/79.0/86.1 \\
& rescue\_min\_step & 8 & 68.0/85.0/80.0 & 75.0/95.0/78.9 & 91.0/99.0/91.9 & 85.0/99.0/85.9 & 64.0/75.0/84.0 \\
& rescue\_min\_step & 12 & 63.0/84.0/75.0 & 72.0/95.0/75.8 & 93.0/100.0/93.0 & 87.0/100.0/87.0 & 63.0/76.0/82.9 \\
& rescue\_min\_step & 16 & 68.0/84.0/81.0 & 73.0/94.0/77.7 & 92.0/100.0/92.0 & 85.0/100.0/85.0 & 62.0/78.0/78.2 \\

\bottomrule
\end{tabular}
\end{table}

\begin{table}[t]
\centering
\scriptsize
\setlength{\tabcolsep}{2.5pt}
\renewcommand{\arraystretch}{1.08}
\caption{Sensitivity to the minimum rescue step on LIBERO-Spatial.
Each cell reports SR/IR/RSR.}
\label{tab:hyper_rescue_spatial}

\begin{tabular}{llcccccc}
\toprule
\textbf{Backbone} & \textbf{Setting} & \textbf{Value}
& \textbf{Task} & \textbf{Pos.} & \textbf{Obj.} & \textbf{Sem.} & \textbf{Env.} \\
\midrule

\multirow{4}{*}{OpenVLA}
& FIT-MAC & -- & 38.0/94.0/38.3 & 41.0/88.0/40.9 & 97.0/34.0/100.0 & 85.0/44.0/97.7 & 99.0/24.0/100.0 \\
& rescue\_min\_step & 8 & 37.0/95.0/36.8 & 39.0/90.0/41.1 & 97.0/31.0/100.0 & 86.0/47.0/97.9 & 97.0/27.0/100.0 \\
& rescue\_min\_step & 12 & 31.0/94.0/30.9 & 41.0/88.0/40.9 & 97.0/31.0/100.0 & 85.0/44.0/97.7 & 99.0/24.0/100.0 \\
& rescue\_min\_step & 16 & 36.0/95.0/35.8 & 39.0/90.0/41.1 & 97.0/31.0/100.0 & 86.0/47.0/97.9 & 97.0/27.0/100.0 \\

\midrule

\multirow{4}{*}{$\pi$0.5}
& FIT-MAC & -- & 25.0/100.0/25.0 & 37.0/98.0/37.8 & 97.0/52.0/100.0 & 93.0/63.0/88.9 & 94.0/41.0/97.6 \\
& rescue\_min\_step & 8 & 36.0/100.0/36.0 & 41.0/90.0/34.4 & 97.0/52.0/98.1 & 95.0/63.0/92.1 & 95.0/42.0/95.2 \\
& rescue\_min\_step & 12 & 39.0/100.0/39.0 & 40.0/90.0/33.3 & 96.0/52.0/96.2 & 95.0/62.0/91.9 & 94.0/42.0/95.2 \\
& rescue\_min\_step & 16 & 41.0/100.0/41.0 & 44.0/90.0/37.8 & 97.0/52.0/98.1 & 94.0/62.0/91.9 & 97.0/42.0/97.6 \\

\bottomrule
\end{tabular}
\end{table}

\subsubsection{Controller-Horizon Sensitivity}
\label{app:hyperparameter_horizon}

The parameter \text{multi-agent-exec-horizon} specifies the maximum number of predicted actions executed before the next policy query and monitoring update in multi-agent control mode. The effective horizon is bounded by the open-loop horizon and given by
\[
H_{\mathrm{effective}}
=
\min\left(
H_{\mathrm{open\text{-}loop}},
H_{\mathrm{exec}}
\right),
\]
where $H_{\mathrm{exec}}$ denotes the configured execution horizon. This parameter controls feedback frequency without changing the action-chunk length predicted by the pretrained VLA.

In Table~\ref{tab:hyper_horizon_object} and Table~\ref{tab:hyper_horizon_spatial}, we report results for horizons of $\{1, 2, 3, 4\}$ on LIBERO-Object and LIBERO-Spatial, respectively. Shorter horizons allow more frequent policy queries and monitoring updates, potentially enabling earlier responses to failures. Longer horizons allow more predicted actions to execute between updates but delay feedback-based correction.

For OpenVLA on LIBERO-Object, increasing the horizon from $1$ to $2$ raises SR under task perturbations from 36.0\% to 86.0\% and under position perturbations from 58.0\% to 94.0\%. A horizon of $3$ further increases SR under task perturbations to 96.0\%, whereas a horizon of $4$ reduces SR under task and position perturbations to 77.0\% and 79.0\%, respectively. On LIBERO-Spatial, however, increasing the horizon from $2$ to $3$ reduces SR under position and environment perturbations from 41.0\% to 39.0\% and from 99.0\% to 97.0\%, respectively.

For $\pi_{0.5}$ on LIBERO-Object, a horizon of $2$ achieves the highest SR under task perturbations at 80.0\%, while a horizon of $4$ performs best under environment perturbations at 81.0\%. On LIBERO-Spatial, increasing the horizon from $1$ to $4$ raises SR under task perturbations from 29.0\% to 38.0\% and under position perturbations from 16.0\% to 46.0\%. These results show that more frequent feedback does not necessarily improve SR and that the preferred execution horizon depends on the backbone and perturbation condition.

\begin{table}[t]
\centering
\scriptsize
\setlength{\tabcolsep}{2.5pt}
\renewcommand{\arraystretch}{1.08}
\caption{Sensitivity to the execution horizon on LIBERO-Object.
Each cell reports SR/IR/RSR.}
\label{tab:hyper_horizon_object}

\begin{tabular}{llcccccc}
\toprule
\textbf{Backbone} & \textbf{Setting} & \textbf{Value}
& \textbf{Task} & \textbf{Pos.} & \textbf{Obj.} & \textbf{Sem.} & \textbf{Env.} \\
\midrule

\multirow{4}{*}{OpenVLA}
& FIT-MAC & -- & 86.0/100.0/86.0 & 94.0/99.0/94.9 & 100.0/100.0/100.0 & 98.0/98.0/100.0 & 49.0/49.0/100.0 \\
& exec\_horizon & 1 & 36.0/100.0/36.0 & 58.0/100.0/58.0 & 100.0/100.0/100.0 & 97.0/98.0/99.0 & 49.0/49.0/100.0 \\
& exec\_horizon & 2 & 86.0/100.0/86.0 & 94.0/100.0/94.0 & 100.0/100.0/100.0 & 98.0/98.0/100.0 & 49.0/49.0/100.0 \\
& exec\_horizon & 3 & 96.0/100.0/96.0 & 94.0/99.0/94.9 & 100.0/100.0/100.0 & 98.0/98.0/100.0 & 49.0/49.0/100.0 \\
& exec\_horizon & 4 & 77.0/100.0/77.0 & 79.0/99.0/79.8 & 99.0/100.0/99.0 & 98.0/98.0/100.0 & 49.0/49.0/100.0 \\

\midrule

\multirow{4}{*}{$\pi$0.5}
& FIT-MAC & -- & 80.0/100.0/80.0 & 67.0/99.0/67.7 & 92.0/100.0/92.0 & 89.0/97.0/91.8 & 69.0/79.0/86.1 \\
& exec\_horizon & 1 & 64.0/100.0/64.0 & 44.0/100.0/44.0 & 88.0/100.0/88.0 & 82.0/92.0/89.1 & 55.0/70.0/78.6 \\
& exec\_horizon & 2 & 80.0/100.0/80.0 & 67.0/99.0/67.7 & 92.0/100.0/92.0 & 89.0/97.0/91.8 & 69.0/79.0/86.1 \\
& exec\_horizon & 3 & 68.0/82.0/82.9 & 68.0/96.0/70.8 & 90.0/100.0/90.0 & 93.0/99.0/93.9 & 67.0/79.0/84.8 \\
& exec\_horizon & 4 & 64.0/85.0/75.3 & 72.0/93.0/77.4 & 89.0/99.0/89.9 & 92.0/100.0/92.0 & 81.0/90.0/90.0 \\

\bottomrule
\end{tabular}
\end{table}

\begin{table}[t]
\centering
\scriptsize
\setlength{\tabcolsep}{2.5pt}
\renewcommand{\arraystretch}{1.08}
\caption{Sensitivity to the execution horizon on LIBERO-Spatial.
Each cell reports SR/IR/RSR.}
\label{tab:hyper_horizon_spatial}

\begin{tabular}{llcccccc}
\toprule
\textbf{Backbone} & \textbf{Setting} & \textbf{Value}
& \textbf{Task} & \textbf{Pos.} & \textbf{Obj.} & \textbf{Sem.} & \textbf{Env.} \\
\midrule

\multirow{4}{*}{OpenVLA}
& FIT-MAC & -- & 38.0/94.0/38.3 & 41.0/88.0/40.9 & 97.0/34.0/100.0 & 85.0/44.0/97.7 & 99.0/24.0/100.0 \\
& exec\_horizon & 1 & 38.0/94.0/38.3 & 40.0/88.0/39.8 & 97.0/34.0/100.0 & 86.0/45.0/97.8 & 99.0/24.0/100.0 \\
& exec\_horizon & 2 & 28.0/94.0/27.7 & 41.0/88.0/40.9 & 97.0/34.0/100.0 & 85.0/44.0/97.7 & 99.0/24.0/100.0 \\
& exec\_horizon & 3 & 31.0/95.0/30.5 & 39.0/90.0/41.1 & 97.0/31.0/100.0 & 85.0/46.0/97.8 & 97.0/27.0/100.0 \\
& exec\_horizon & 4 & 29.0/95.0/28.4 & 39.0/90.0/41.1 & 97.0/31.0/100.0 & 84.0/46.0/95.7 & 97.0/27.0/100.0 \\

\midrule

\multirow{4}{*}{$\pi$0.5}
& FIT-MAC & -- & 25.0/100.0/25.0 & 37.0/98.0/37.8 & 97.0/52.0/100.0 & 93.0/63.0/88.9 & 94.0/41.0/97.6 \\
& exec\_horizon & 1 & 29.0/100.0/29.0 & 16.0/98.0/14.3 & 98.0/52.0/100.0 & 86.0/60.0/80.0 & 93.0/40.0/97.5 \\
& exec\_horizon & 2 & 25.0/100.0/25.0 & 37.0/98.0/37.8 & 97.0/52.0/100.0 & 93.0/63.0/88.9 & 94.0/41.0/97.6 \\
& exec\_horizon & 3 & 35.0/100.0/35.0 & 45.0/98.0/43.9 & 95.0/52.0/94.2 & 91.0/63.0/87.3 & 98.0/42.0/100.0 \\
& exec\_horizon & 4 & 38.0/100.0/38.0 & 46.0/97.0/44.3 & 98.0/52.0/100.0 & 92.0/62.0/88.7 & 93.0/42.0/92.9 \\

\bottomrule
\end{tabular}
\end{table}

\section{Additional Qualitative Results}
\label{app:qualitative_results}

We present additional qualitative examples from LIBERO-PRO to illustrate how FIT-MAC addresses target-binding errors and local execution failures. Each example shows the robot state before and after intervention, highlighting the initial error, corrective action, and subsequent behavior. Red annotations mark the incorrect target or destination, while green annotations indicate the corrected target or destination. These examples complement the quantitative results by illustrating how targeted corrections guide the robot's actions during execution.

\subsection{Target-Correction Cases}
\label{app:target_examples}

\begin{figure}[t]
    \centering
    \includegraphics[
        width=\textwidth
    ]{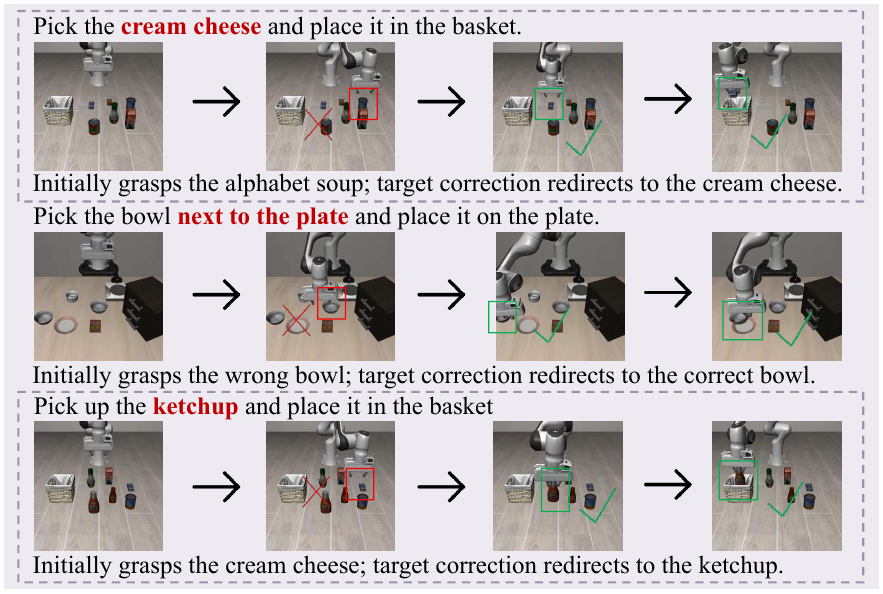}
    \vspace{-18pt}
\caption{Qualitative target-correction cases. In each row, the policy initially selects or approaches an incorrect object (red). FIT-MAC detects the error, corrects the target binding, and redirects manipulation toward the intended object (green).}
    \label{fig:target_correction_examples}
\end{figure}

In Figure~\ref{fig:target_correction_examples}, we present representative cases in which the policy initially approaches or grasps an object that does not match the instruction. These target-binding errors concern object selection rather than manipulation execution. FIT-MAC detects the mismatch, corrects the target binding, and redirects subsequent actions toward the intended object. The states before and after intervention illustrate how target correction changes the robot's immediate behavior and aligns object selection with the task instruction.

\subsection{Execution-Correction Cases}
\label{app:execution_examples}

\begin{figure}[t]
    \centering
    \includegraphics[width=\textwidth]{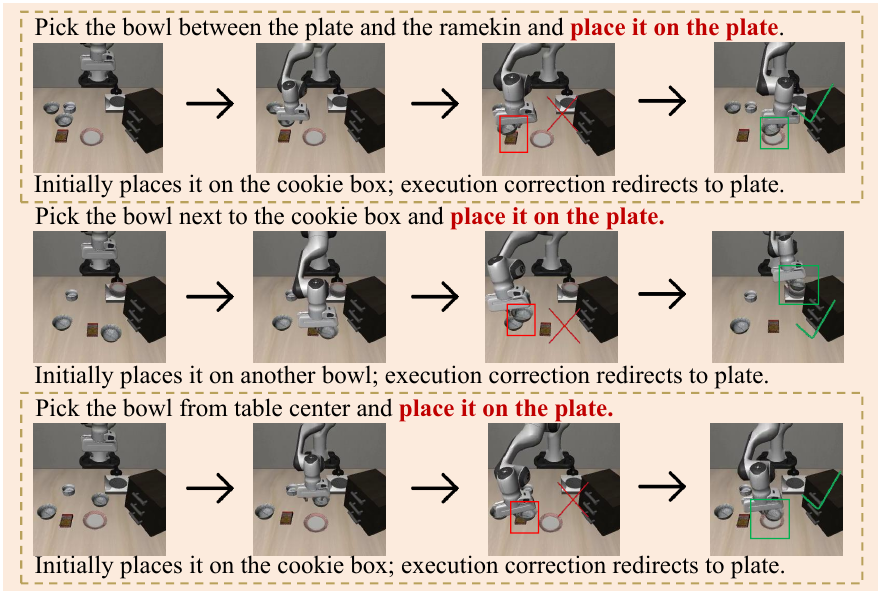}
    \vspace{-18pt}
    \caption{Qualitative execution-correction cases. The robot initially transports or places the selected bowl at an incorrect destination (red). FIT-MAC detects the placement error and redirects the bowl to the instructed plate (green).}
    \label{fig:execution_correction_examples}
\end{figure}

In Figure~\ref{fig:execution_correction_examples}, we present cases in which the policy selects the correct object but transports it to or places it at the wrong destination. For example, the robot may place the object on a cookie box or in another bowl rather than on the specified plate. Although object selection is correct, the resulting placement violates the task requirement. FIT-MAC detects the placement error and invokes local recovery to move the object to the intended destination. These sequences illustrate how local execution correction resolves placement errors while preserving the target binding.

\subsection{Failure Cases}
\label{app:failure_cases}

\begin{figure}[t]
    \centering
    \includegraphics[
        width=\textwidth
    ]{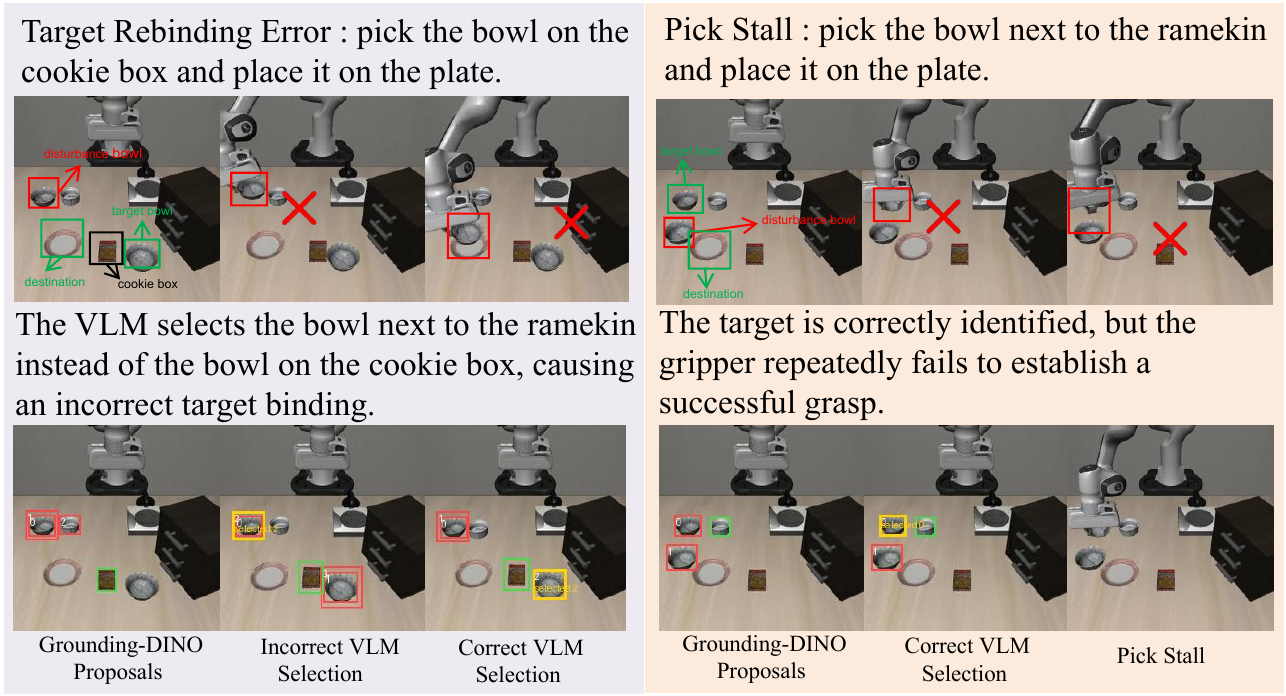}
    \vspace{-18pt}
    \caption{Representative FIT-MAC failures.
    Left: target rebinding error,
    The VLM selects the bowl next to the ramekin rather than the instructed bowl, causing the controller to grasp the wrong object.
    Right: pick stall,
    The target is correctly identified, but repeated grasp attempts fail.}
    \label{fig:wrongcase}
\end{figure}

In Figure~\ref{fig:wrongcase}, we illustrate two failure modes involving incorrect target rebinding and stalled grasping despite correct target identification. In the left example, the robot is instructed to pick the bowl next to the cookie box and place it on the plate. Grounding DINO detects candidate bowls, but the VLM selects the bowl next to the ramekin, violating the spatial relation specified in the instruction. The controller follows this incorrect binding and grasps the wrong bowl. This failure originates in target selection rather than low-level control. Completing the grasp therefore does not resolve the upstream grounding error.

In the right example, the robot is instructed to pick the bowl next to the ramekin and place it on the plate. The VLM correctly identifies the target among the candidates detected by Grounding DINO, but the gripper repeatedly fails to grasp it after approaching. These failed attempts stall execution and prevent progression to placement. This case shows that correct grounding alone does not ensure successful manipulation. Together, these examples highlight remaining limitations in spatial grounding and grasp execution, either of which can prevent task completion.

\section{FIT-MAC Execution and Implementation}
\label{app:implementation}

\subsection{FIT-MAC Execution Algorithm}
\label{app:algorithm}

FIT-MAC augments a frozen VLA policy with a runtime controller for grounding, failure diagnosis, target correction, and execution correction. Target correction revises the selected object or destination before action generation. Execution correction addresses failures during manipulation, including unsuccessful grasping, failed lifting, stalled transport, and incorrect placement. The controller selects the appropriate response at policy-query boundaries without modifying the underlying VLA parameters.

Let $\mathcal{P}$ denote the ordered subtask plan, $G$ the persistent grounding record, $S$ the attempt-monitoring state, $R$ the persistent execution-correction state, and $\mathcal{H}$ the diagnostic history. The grounding record stores the selected target, confidence, target-instance lock, image location, and destination reference when available. Once a target instance is resolved, subsequent monitoring tracks that instance rather than independently selecting a candidate at each query.

At each policy-query boundary, FIT-MAC identifies the active subtask and initializes its grounding state when needed. The Verifier Agent summarizes the current manipulation attempt and diagnoses the observed evidence. The adaptive gate then determines whether to permit target or execution correction based on the diagnosis, target confidence, instance reliability, execution progress, and failure persistence. Pending execution correction takes precedence in routing. When no correction is active or permitted, the pretrained VLA generates actions using the current grounded instruction.

\begin{algorithm}[t]
\caption{FIT-MAC runtime execution}
\label{alg:fit_mac}
\small
\begin{algorithmic}[1]
\Require Instruction $x$, pretrained policy $\pi$, environment $\mathcal{E}$,
configuration $\theta$, execution budget $T$
\Ensure Verified outcome $y$ and episode trace $\mathcal{L}$

\State $\mathcal{P} \gets \Call{ParseAndDecompose}{x}$
\State Initialize grounding record $G$, monitoring state $S$, and correction state $R$
\State Initialize verifier $V$, diagnostic history $\mathcal{H}$,
action queue $Q$, and trace $\mathcal{L}$
\State $y \gets \Call{UnknownOutcome}{}$

\For{$t=0,\ldots,T-1$}
    \If{$Q=\emptyset$}
        \State $o_t \gets \Call{Observe}{\mathcal{E}}$
        \State $j \gets \Call{SelectActiveSubtask}{\mathcal{P},V,G}$
        \State $x_j \gets \mathcal{P}[j]$

        \If{$x_j$ is new or $G$ is not initialized}
            \State $G \gets
            \Call{GroundTargetAndDestination}{x_j,o_t,\theta}$
        \EndIf

        \State $S \gets
        \Call{MonitorAttempt}{S,R,G,o_t,V,\mathcal{E}}$
        \State $(D,\mathcal{H}) \gets
        \Call{Diagnose}{S,\mathcal{H},G,\theta}$
        \State $\Gamma \gets
        \Call{AdaptiveGate}{D,S,G,\mathcal{H},\theta}$

        \If{\Call{HasPendingExecutionCorrection}{R}}
            \State $M \gets \textnormal{pending execution correction}$
        \ElsIf{$\Gamma.\textnormal{allow\_target\_correction}$}
            \State $M \gets \textnormal{target correction}$
        \ElsIf{$\Gamma.\textnormal{allow\_execution\_correction}$}
            \State $M \gets \textnormal{execution correction}$
        \Else
            \State $M \gets \textnormal{base policy}$
        \EndIf

        \If{$M=\textnormal{target correction}$}
            \State $G \gets
            \Call{CorrectTargetGrounding}{G,S,o_t,\theta}$
            \State $\ell \gets
            \Call{BuildPolicyInstruction}{x_j,G,\theta}$
            \State $A \gets \pi(\ell,o_t)$

        \ElsIf{$M\in
        \{\textnormal{pending execution correction},
        \textnormal{execution correction}\}$}
            \State $(A,R) \gets
            \Call{RunExecutionCorrection}{R,G,o_t,S,V,\mathcal{E},\theta}$

            \If{$A=\bot$}
                \State $\ell \gets
                \Call{BuildPolicyInstruction}{x_j,G,\theta}$
                \State $A \gets \pi(\ell,o_t)$
            \EndIf

        \Else
            \State $\ell \gets
            \Call{BuildPolicyInstruction}{x_j,G,\theta}$
            \State $A \gets \pi(\ell,o_t)$
        \EndIf

        \State $H_{\mathrm{eff}} \gets
        \Call{EffectiveHorizon}{\theta,\Gamma}$
        \State $Q \gets
        \Call{Prefix}{A,\min(H_{\mathrm{eff}},|A|)}$
        \State Append the routing decision and controller state to $\mathcal{L}$
    \EndIf

    \State $a_t \gets \Call{PopFront}{Q}$
    \State $(o_{t+1},\mathrm{terminal}) \gets
    \Call{Step}{\mathcal{E},a_t}$
    \State Append $(a_t,o_{t+1})$ to $\mathcal{L}$

    \If{$\mathrm{terminal}$}
        \State \textbf{break}
    \EndIf
\EndFor

\State $y \gets
\Call{VerifyEpisodeOutcome}{\mathcal{E},V,\mathcal{L}}$
\State \Return $(y,\mathcal{L})$
\end{algorithmic}
\end{algorithm}

Here, $\bot$ indicates that the execution-correction controller provides no valid action for the current query, causing control to return to the pretrained VLA. In multi-agent control mode, the effective execution horizon is
\begin{equation}
H_{\mathrm{eff}}
=
\min\left(
H_{\mathrm{open\text{-}loop}},
H_{\mathrm{exec}}
\right),
\end{equation}
where $H_{\mathrm{exec}}$ denotes the configured value of \text{multi-agent-exec-horizon}. This horizon limits the number of predicted actions executed before the next policy query and monitoring update. It does not change the action-chunk length predicted by the pretrained VLA.

\noindent\textbf{Grounding and Persistent Target State.}
For each active subtask, the Grounder Agent identifies the target object from the instruction and current observation. Candidate detections are filtered by object category and spatial description. When detections are missing, unreliable, or ambiguous, an additional language-model query may be used within the configured query budget. The grounding record stores the selected target, confidence evidence, target-instance lock, image location, and destination reference when available.
The target binding persists across policy queries unless explicitly corrected, allowing monitoring and correction to track a consistent object instance. When the active subtask changes, the grounding state is reinitialized for the new instruction.

\noindent\textbf{Runtime Verification.}
The Verifier Agent checks target consistency and execution progress using gripper state, end-effector motion, object lift, and target--gripper geometry. Phase-specific criteria identify wrong-target approaches, unsuccessful grasps or lifts, stalled transport, and failed placement.
Using the current execution state and recent verification history, the Verifier Agent distinguishes target-binding errors from local execution failures. Target-binding errors indicate an incorrect or unreliable target binding or destination reference. Local execution failures indicate a lack of expected manipulation progress. An optional model assessment supplements these checks, and the supporting evidence and verification results are recorded in the runtime trace.

\noindent\textbf{Adaptive Intervention Gate.}
Failure detection does not automatically trigger correction. The Verifier Agent uses an adaptive gate to determine whether the available evidence is sufficiently reliable and persistent to justify intervention.
For object, semantic, and environment perturbation routes, the gate applies an additional progress guard. When the pretrained VLA is making recognized progress, execution continues unless stronger evidence of a target-binding error or local execution failure is available. Failure diagnosis and correction activation are therefore separate decisions. FIT-MAC can record a failure while continuing to monitor the pretrained VLA when the evidence is insufficient to justify correction.

\noindent\textbf{Target Correction.}
Target correction updates the grounding state used to construct the policy instruction. When the Verifier Agent identifies a wrong-target approach, unreliable target lock, ambiguous target instance, or incorrect destination reference, the Correction Agent invokes the Grounder Agent. The Grounder Agent revisits the available evidence, updates the target binding or destination representation, and rebuilds the grounded instruction.
This correction changes the object or destination specified to the policy without replacing its low-level action generation. The pretrained VLA then generates the next action chunk using the revised grounding record.

\noindent\textbf{Execution Correction.}
Execution correction addresses failures after manipulation begins and maintains persistent phase-specific controller state across policy queries. The Correction Agent selects a recovery routine according to the current manipulation phase and available failure evidence.

The pick-correction routine approaches the selected target, descends, closes the gripper, and verifies whether the object has been lifted. If lift verification fails, the controller can reopen the gripper and initiate another controlled descent. Once carrying is confirmed, control passes to the placement routine.
The place-correction routine verifies that the object is being carried and resolves a valid destination anchor. It then lifts the object, transports it above the destination, adjusts the object--destination alignment, descends, releases the object, and retreats. Geometric placement control applies only when the destination can be represented by a direct anchor. Instructions requiring more complex destination relations remain under pretrained VLA control.

After release, an optional settling routine keeps the gripper open and applies a small upward motion for a configured number of queries. This step reduces the risk of continued gripper contact or immediate disturbance after release. Execution correction is cancelled if its validity conditions fail, the current attempt enters an incompatible phase, or the correction query budget is exhausted. In each case, the controller returns $\bot$, and control reverts to the pretrained VLA.

\vspace{-5pt}

\subsection{Representative Runtime Traces}
\label{app:runtime_traces}

In Table~\ref{tab:runtime_trace_spatial}, we present an event-level trace of a successful LIBERO-Spatial episode under Task perturbation. The instruction is \emph{``Pick the akita black bowl not between the plate and the ramekin and place it on the plate.''} Two black bowls are detected in the initial observation, requiring the Grounder Agent to resolve the specified spatial relation before execution.

During target correction, the Grounder Agent selects the intended object from the detected candidates based on the instruction, using VLM-based candidate selection when needed. It then establishes a persistent target-instance lock. During execution correction, the Correction Agent applies phase-specific control to recover from failures in approach, grasping, lifting, transport, alignment, or placement.

\begin{table}[t]
\centering
\small
\setlength{\tabcolsep}{4pt}
\renewcommand{\arraystretch}{1.16}

\caption{Representative runtime trace of a successful LIBERO-Spatial episode under Task perturbation. Target correction resolves ambiguity between candidate bowls during grounding. Execution correction activates after a wrong-target approach and remains active through placement correction.}
\label{tab:runtime_trace_spatial}

\begin{tabularx}{\textwidth}{
    @{}
    >{\centering\arraybackslash}p{0.055\textwidth}
    >{\centering\arraybackslash}p{0.050\textwidth}
    >{\raggedright\arraybackslash}p{0.095\textwidth}
    >{\raggedright\arraybackslash}p{0.255\textwidth}
    >{\raggedright\arraybackslash}p{0.135\textwidth}
    >{\raggedright\arraybackslash}X
    @{}
}
\toprule
\textbf{Query}
& \textbf{Step}
& \textbf{Phase}
& \textbf{Event / evidence}
& \textbf{Diagnosis and route}
& \textbf{Controller state} \\
\midrule

1
& 10
& Pick approach
& Two bowl candidates are detected. Grounding resolves the spatial relation and selects the black bowl[1] with VLM confidence $0.95$.
& Target grounding; target correction completed.
& Target-instance lock established; base policy starts execution. \\

5
& 42
& Pick approach
& The robot continues approaching the resolved target.
& None; baseline execution.
& Base policy continues with an eight-action chunk. \\

6
& 50
& Pick approach
& The monitor reports a wrong-target approach.
& Execution failure; execution correction activated.
& Hard-pick correction starts in approach-hover phase. \\

7
& 51
& Pick approach
& The correction state remains pending.
& Execution failure; correction remains active.
& Hard-pick controller continues with one corrective action. \\

30
& 74
& Place transport
& The pick attempt transitions successfully to placement. The object is reported as lifted and carried.
& Successful pick-to-place transition.
& Destination-locked place correction starts in lift-carry phase. \\

31
& 75
& Place transport
& The monitor reports insufficient placement progress and a placement stall.
& Execution failure; correction remains active.
& Destination is locked to the plate\_1. \\

33
& 77
& Place transport
& The object has not reached the destination.
& Execution failure; correction remains active.
& Move-above-destination phase performs alignment. \\

55
& 99
& Place transport
& The controller progresses to the descent stage.
& Execution failure; correction remains active.
& Descent-to-destination phase; release height not yet reached. \\

59
& 103
& Place descend
& The object reaches the release condition.
& Execution correction; release permitted.
& Release phase enters the release-ready state. \\

60
& 104
& Place release
& The controller holds the release state while waiting for the gripper to open.
& No new failure; correction remains pending.
& Release-hold phase. \\

61
& 105
& Place completion
& The environment returns reward $1$ and terminates successfully.
& Successful completion.
& Episode ends after the final release-hold action. \\

\bottomrule
\end{tabularx}
\end{table}

\subsection{Agent Prompts and Interfaces}
\label{app:agent_prompts}

We document the prompts and interfaces of FIT-MAC's four agents. The Planner Agent converts the instruction into a structured task specification, and the Grounder Agent resolves the target instance and destination. The Verifier Agent monitors execution, diagnoses failures, and determines whether the available evidence justifies intervention. The Correction Agent routes permitted interventions to target rebinding or bounded local execution recovery. We provide the available prompts and describe how their outputs support grounding, verification, and correction.

\subsubsection{Planner Agent}
\label{app:task_planner_prompt}

The Planner Agent converts the natural-language LIBERO instruction into a canonical instruction and a coarse-grained execution plan. It preserves the instruction's granularity rather than splitting a simple pick-and-place task into separate locate, grasp, move, and release subtasks. A rule-based planner provides a fallback, while an optional local language model produces the same structured interface.

The system prompt is provided below.

\begin{promptbox}[title={Planner Agent System Prompt}]
You are a task decomposition agent.
Given a natural-language manipulation instruction, produce a structured task description.
Output JSON only with the following keys:
canonical_instruction
plan
action
target_object
source_phrase
destination
destination_prep
destination_relation
confidence
Keep the plan at the same granularity as the instruction.
Preserve object identities, spatial relations, and destination references.
Return only valid JSON.
\end{promptbox}

\subsubsection{Grounder Agent}
\label{app:grounder_prompt}

The Grounder Agent obtains candidate objects from Grounding DINO and interprets the target description and spatial relations in the task specification. It invokes a VLM candidate selector when the candidates are ambiguous or detection confidence is insufficient. The resulting grounding record establishes a target-instance lock that persists during manipulation unless explicitly corrected.

For candidate selection, the system prompt is provided below.

\begin{promptbox}[title={Grounder Agent System Prompt}]
You are a grounding agent.
Follow the instructions to select the correct target object.
The green-labeled box is a reference object when present.
The red-labeled boxes are the only candidate target objects that may be selected.
Do not select the reference object, the robot arm, or any other non-target object.
Output JSON only with the following keys:
selected_index, confidence, reason
Use zero-based indexing. Choose the target by the source phrase and its
spatial relation to the reference object. Do not choose by candidate order or detector score alone.
\end{promptbox}

The corresponding user prompt supplies the instruction, reference objects, and detected candidates.

\begin{promptbox}[title={Grounder Agent User Prompt}]
Instruction: <instruction>
Canonical instruction: <canonical_instruction>
Source phrase: <source_phrase>
Target object: <target_object>
Reference objects: <reference_objects>
Candidates:
<candidate_list>
Other detected instances:
<other_instances>
Selection rules:
- Select only a red candidate box.
- Use the source phrase and its relation to the reference object.
- Ignore candidate order unless multiple candidates are equally plausible.
- Do not select a reference object or an extra detection.
- Return only valid JSON.
\end{promptbox}

\subsubsection{Verifier Agent}
\label{app:monitor_diagnosis_prompt}

The Verifier Agent combines execution monitoring and failure diagnosis. It tracks manipulation progress using deterministic state and motion signals without replacing the pretrained VLA's action generation. Its diagnosis interface distinguishes no actionable failure, target failure, execution failure, and task success.

An optional language-model assessment supplements the rule-based diagnosis. Its output is normalized, its confidence is clamped, and its evidence is merged with the deterministic checks before intervention gating and correction routing.

The Verifier Agent first checks task completion and then determines whether an actionable failure is present. If no attempt-end condition is present, it returns no actionable failure. If correction is already pending, it records an intermediate status indicating that recovery remains in progress.

Evidence for target-binding errors includes target ambiguity, target-probe mismatch, and inconsistency between the selected target and the instruction. Evidence for local execution failures includes phase stalls, unsuccessful grasps or lifts, failed placement, and lack of manipulation progress. A wrong-target approach during execution is treated as an execution failure when the target binding remains correct.

When enabled, the optional language-model assessment uses the following system prompt.

\begin{promptbox}[title={Verifier Agent System Prompt}]
You are a critic agent.
Inspect the task instruction and the execution snapshot. Output JSON only with the following keys:
failure_type
reason
confidence
evidence_tags
failure_type must be one of:
none, target_failure, execution_failure, success.
Use target_failure when the target object is likely wrong, ambiguous, or
mismatched with the instruction. Use execution_failure when the target seems
correct but the robot stalled, failed to grasp, failed to place, or made no
progress. Use none when there is no actionable failure. Keep the reason
short and factual.
\end{promptbox}

The corresponding user prompt provides the active instruction and execution snapshot.

\begin{promptbox}[title={Verifier Agent User Prompt}]
Task label: <active_instruction>
Snapshot:
{
  "route": "<route>",
  "phase": "<phase>",
  "attempt_kind": "<attempt_kind>",
  "attempt_id": <attempt_id>,
  "phase_steps": <phase_steps>,
  "stall_reason": "<stall_reason>",
  "attempt_end_candidate": <true_or_false>,
  "gripper_closed": <true_or_false>,
  "target_lifted": <true_or_false>,
  "target_selected": "<target_selected>",
  "target_confidence": <target_confidence>,
  "progress_flags": <progress_flags>,
  "verifier_status": "<verifier_status>",
  "target_probe": <target_probe>,
  "rule_hint": "<rule_hint>",
  "recent_contexts": <recent_contexts>,
  "recent_diagnoses": <recent_diagnoses>
}
Return only valid JSON.
\end{promptbox}

The failure labels represent diagnosis categories rather than direct intervention commands. Target failures correspond to target-binding errors, while execution failures are further distinguished by manipulation phase for recovery routing. Correction activation additionally depends on the adaptive gate and any pending execution correction.

\subsubsection{Correction Agent}
\label{app:correction_interface}

The Correction Agent uses the Verifier Agent's diagnosis and intervention permissions, together with the shared execution state, to select a corrective response. We summarize its input and output interface below.

\begin{promptbox}[title={Correction Agent Input and Output}]
Input
- Failure diagnosis and supporting evidence
- Adaptive intervention-gate permissions
- Active subtask and manipulation phase
- Current observation and grounding record
- Persistent execution-correction state
- Recovery budget and controller configuration

Output
- Selected route: pretrained VLA, target correction, or execution correction
- For target correction: updated grounding record and grounded instruction
- For execution correction: corrective actions and updated correction state
- If no valid corrective action is available: fallback signal
\end{promptbox}

For target correction, the agent obtains the updated grounding record and instruction through the Grounder Agent. For execution correction, it invokes the appropriate bounded local controller. Pending execution correction takes precedence, and a fallback signal returns control to the pretrained VLA.

\end{document}